\pdfoutput=1

\newif\ifconference
\newif\iffull

\conferencetrue
\fullfalse

\documentclass[sigconf]{acmart}

\usepackage[utf8]{inputenc}
\usepackage[english]{babel}

\usepackage{amsfonts,amstext,amsthm}
\usepackage{dashbox}
\usepackage{booktabs}
\usepackage{color}
\definecolor{dkgreen}{rgb}{0,0.6,0}
\definecolor{gray}{rgb}{0.5,0.5,0.5}
\definecolor{mauve}{rgb}{0.58,0,0.82}
\usepackage{listings}
\usepackage{rotating}
\usepackage{caption,subcaption}
\usepackage{pifont}
\usepackage{stmaryrd}
\usepackage{array}
\usepackage[]{algorithm2e}
\usepackage{graphicx}
\usepackage{mathtools}
\usepackage{adjustbox}
\usepackage{float}
\usepackage{multirow}
\usepackage{makecell}
\usepackage{tikz}
\usetikzlibrary{arrows.meta,decorations.pathmorphing,decorations.footprints,fadings,calc,trees,mindmap,shadows,decorations.text,patterns,positioning,shapes,matrix,fit}
\usepackage{pgfplots}
\pgfplotsset{compat=1.14}
\usepackage{url}
\usepackage{varioref}
\usepackage{enumitem}

\newcommand{\new}[1]{\textcolor{black}{#1}}

\newcommand{\ie}{{i.e.,~}}
\newcommand{\loss}{\mathcal{L}}
\newcommand\myeq{\mkern1.5mu{=}\mkern1.5mu}
\usepackage[
	operators,
	sets,
	adversary,
	landau,
	probability,
	notions,
	logic,
	ff,
	mm,
	advantage,
	primitives,
	events,
	complexity,
	asymptotics,
	keys,
	lambda]{cryptocode}
\usepackage{hyperref}
\newcommand\myshade{85}
\colorlet{mylinkcolor}{violet}
\colorlet{mycitecolor}{orange}
\colorlet{myurlcolor}{blue}
\hypersetup{
  linkcolor  = mylinkcolor!\myshade!black,
  citecolor  = mycitecolor!\myshade!black,
  urlcolor   = myurlcolor!\myshade!black,
  colorlinks = true,
}
\usepackage{cleveref}
\usepackage{tabularx}

\newcounter{protocolenv}
\newenvironment{protocolenv}[1]
  {\par\addvspace{\topsep}
   \noindent
   \tabularx{\linewidth}{@{} X @{}}
    \hline
    \refstepcounter{protocolenv} \textbf{#1} \\
    \hline}
  { \\
   \endtabularx
   \par\addvspace{\topsep}}

\newcommand{\sbline}{\\[.5\normalbaselineskip]}% small blank line

\theoremstyle{definition}

\newcommand{\getsr}{{\:{\leftarrow{\hspace*{-3pt}\raisebox{.75pt}{$\scriptscriptstyle\$$}}}\:}}

\newcommand{\relu}{ReLU}
\newcommand{\TT}[1]{``\textit{#1}''}
\newcommand{\eg}{{e.g.,~}}
\newcommand{\etal}{{et~al.}}
\newcommand{\cUsers}{\mathcal{U}}

\newcommand{\cX}{\mathcal{X}}

\newcommand{\cY}{\mathcal{Y}}

\newcommand{\dataset}{\mathcal{D}}

\newcommand{\labeltarget}{\mathsf{trgt}}

\newcommand{\labelSecAggr}{\mathsf{sa}}

\setcopyright{none} % No copyright notice required for submissions
\acmConference[Anonymous Submission to ACM CCS 2022]{ACM Conference on Computer and Communications Security}{Due 15 May 2019}{London, TBD}
\acmYear{2019}

\begin{document}

\title[Eluding Secure Aggregation in Federated Learning via Model Inconsistency]{Eluding Secure Aggregation in Federated Learning \\ via Model Inconsistency}\titlenote{In the proceedings of ACM Conference on Computer and Communications Security 2022 (CCS '22).}

\author{Dario Pasquini}
\email{dario.pasquini@epfl.ch}
\affiliation{%
  \institution{SPRING Lab, EPFL}
  %\streetaddress{P.O. Box 1212}
  \city{Lausanne}
  \country{Switzerland}
}

\author{Danilo Francati}
\email{dfrancati@cs.au.dk}
\affiliation{%
  \institution{Aarhus University}
  \city{Aarhus}
  \country{Denmark}
 }

\author{Giuseppe Ateniese}
\email{ateniese@gmu.edu}
\affiliation{%
  \institution{George Mason University}
  \city{Fairfax, Virginia}
  \country{USA}
}

\begin{abstract}

Secure aggregation is a cryptographic protocol that securely computes the aggregation of its inputs. It is pivotal in keeping model updates private in federated learning. Indeed, the use of secure aggregation prevents the server from learning the value and the source of the individual model updates provided by the users, hampering inference and data attribution attacks.

In this work, we show that a malicious server can easily elude secure aggregation as if the latter were not in place.
We devise two different attacks capable of inferring information on individual private training datasets, independently of the number of users participating in the secure aggregation.
This makes them concrete threats in large-scale, real-world federated learning applications.

The attacks are generic and equally effective regardless of the secure aggregation protocol used.
They exploit a vulnerability of the federated learning protocol caused by incorrect usage of secure aggregation and lack of parameter validation. Our work demonstrates that current implementations of federated learning with secure aggregation offer only a ``false sense of security''.
% , including the most influential and practical secure aggregation protocols proposed by Bonawitz et al. (CCS '17) and Bell et al. (CCS '20).

  \end{abstract}


\maketitle

\section{Introduction}\label{sec:intro}
Deep learning is evolving rapidly but often at the expense of privacy and security. Neural networks may misbehave, hide backdoors, or be reverse-engineered to reveal sensitive information about the training datasets~\cite{ateniese2015hacking, shokri2017membership, fredrikson2015model}.
Data holders are thus reluctant to provide and share their datasets unless some level of protection is in place.

Cryptographic primitives, such as multi-party computation (MPC) and fully homomorphic encryption (FHE), offer only a partial solution to this problem: They enable learning while protecting sensitive information but at the expense of efficiency and scalability. Even state-of-the-art implementations of these primitives are highly inefficient and add a significant overhead to the learning process, making them unusable and inapplicable in practice.

Accordingly, researchers have looked at alternative solutions that rely on decentralization, where data remain local with the participants while the neural network evolves during the distributed learning process.
Along this line of research, \textbf{federated learning} (FL)~\cite{federated0,federated1,federated2}, along with its main implementations federated stochastic gradient descent (FedSGD) and federated averaging (FedAVG), has been proposed. At a high level, FL allows a set of users to train a shared neural network without outsourcing their local datasets.
To this end, they are only required to locally train the neural network and send model updates (e.g., gradients, model parameters) to a central server.
The updates will be aggregated by the server, completing a round of the training.
The informal security guarantee offered by FL is that sharing the (possibly scrambled~\cite{dp}) updates does not leak any information about the actual training instances used by the users.
Unfortunately, it has been shown that an adversary can invert an individual model update of a target user in order to leak a large amount of information about its dataset~\cite{hitaj2017deep, 8835245, invg1, melis2019exploiting}.\par

For this reason, Bonawitz et al.~\cite{bonawitz2017practical} have proposed to combine \textbf{secure aggregation} (SA) protocols with FL as a first step to increase the security of FL, preventing the server from accessing individual model updates.
Informally, SA is a specialized MPC protocol that allows a set of users to compute the sum of their private inputs securely.
The security guarantee is the same as standard MPC protocols, i.e., nothing is leaked about the inputs except what can be inferred from the output (the sum of the values).

 SA is believed to be one of the most robust defenses against gradient inversion and related inference attacks~\cite{huang2021evaluating}. In particular, the application of SA in FL has two main objectives:
\textbf{(1)}~\TT{Privacy by aggregation}:~Aggregating together a suitable number of model updates smooths out the information carried out by individual contributions. In turn, this makes it unfeasible to assert or recover meaningful information on individual training instances that produced the aggregated value.
\textbf{(2)}~\TT{Privacy by shuffling}:~SA ``hides'' the source of the aggregated information; even if sensitive data is recovered from the aggregated model updates, this cannot be attributed to the user (the individual model update) who provided it.
Thus, although the privacy of the set of users may be violated, the privacy of individuals is preserved.

Our work shows that a motivated and malicious server can easily violate both of these fundamental properties of current SA defenses.
This vulnerability emerges from the federated learning protocol, not the SA protocol; it is caused by incorrect usage of the SA protocol.
For example, even if we abstract the SA protocol with an ideal aggregation functionality, the protocol is still exploitable. In this case, the failure to validate SA inputs by the user is one of the protocol's weaknesses, and we are not targeting a specific implementation of SA.
The main intuition is that model updates (i.e., the inputs of SA) are under the indirect control of the malicious server since model updates are computed starting from the parameters sent by the server.
A malicious server can leverage this control to tamper with the updates (that are the inputs of SA) so that their aggregation will leak information about the update of a target user.

In order to achieve this, the server exploits a new attack vector that we call \textbf{model inconsistency}. Here, the server distributes different views of the same model to different users within the same round.
In this work, we show that model inconsistency can introduce new vulnerabilities in FL algorithms.
The intuition is that a malicious server, providing different parameters to different users, can exploit behavioral differences in the model updates provided by the different models to infer information on users' datasets, even when those model updates are securely aggregated before reaching the server.

To make this inherent vulnerability evident, we implement two attacks that give a representative view of the threat induced by the model inconsistency attack vector.
Eventually, these attacks demonstrate how a malicious server can nullify the security offered by current {SA-based} defenses proposed for FL. That is: \textbf{(1)}~individual model updates can be perfectly recovered from the final aggregated value, independently of the number of users participating in the aggregation, and, \textbf{(2)}~the source of the recovered data can be attributed to individuals in the pool of active users.

\new{Finally, we introduce multiple strategies for preventing the vulnerability that was brought up.
%While existing local differential privacy based approaches can thwart the attacks, 
The proposed solutions seamlessly integrate with current state-of-the-art SA protocols without impacting performance or utility.}%that have relevance in practice.

\subsection{Contributions}\label{subsec:contribution}
Our contributions can be summarized as follows:

\paragraph{Model inconsistency}\label{sec:contribution-model-inconsistency}
We demonstrate that FL incorrectly leverages SA~\cite{bonawitz2017practical}, and as a result, it is as secure as the original FL protocol (without SA).
We prove this by introducing a new adversarial strategy, named \textbf{model inconsistency}, that leverages the following two observations: $(i)$ in each protocol round, the SA's input of user $u_i \in \cUsers$ is its model update $\Delta^{\Theta}_{\dataset_i}$ and, $(ii)$ the value of $\Delta^{\Theta}_{\dataset_i}$ of each $u_i\in \cUsers$ depends on the parameters $\Theta$ sent by the server $S$
(i.e., different parameters produce different model updates).
The combination of the above two observations implies that $S$ could act maliciously and craft different parameters for different users
in order to tamper with the inputs of SA.
As our two attacks will demonstrate, at a different scale, $S$ can exclude from the aggregation the updates of some non-target users $\cUsers \setminus \{u_\labeltarget\}$, forcing the SA to leak part of the model update $\Delta^{\Theta}_{\dataset_\labeltarget}$ of the target~$u_\labeltarget$.

\paragraph{Gradient suppression attack}\label{sec:contribution-gradient-supression}
In~\Cref{sec:attack1}, we present a first attack, named \textbf{gradient suppression}.
It shows that a malicious server can force the local training of the deep model $f_{\widetilde{\Theta}}$ executed by $u_i$ to unconditionally produce a zeroed gradient $\Delta^{\widetilde{\Theta}}_{\dataset_i} \myeq [0]$.
By combining both gradient suppression and model inconsistency, the server $S$ can send the honest parameters $\Theta$ to the target user $u_\labeltarget$ and the malicious parameters $\widetilde{\Theta}$ to the remaining non-target ones $\cUsers\setminus\{u_\labeltarget\}$.
In turn, this will leak the honest $u_\labeltarget$'s gradient $\Delta^{\Theta}_{\dataset_\labeltarget}$ even if SA is in place.
This is because SA will sum up the zeroed gradients of $\cUsers\setminus\{u_\labeltarget\}$ and the honest gradient of $u_\labeltarget$.
The output will be equal to the gradient $\Delta^{\Theta}_{\dataset_\labeltarget}$ of the latter user $u_\labeltarget$.
This is the first practical attack that demonstrates that a malicious server can completely nullify SA. More importantly, the attack does not require auxiliary information on the targets, e.g., the distribution of users' datasets, or unrealistic architecture alterations.

\paragraph{Canary-gradient attack}
We extend the first attack and devise a second approach called \textbf{canary-gradient}. Here, we show that a malicious server $S$ can modify the target's model to induce specific behavior in the derivative of a tiny subset $\xi$ of its parameters (\eg two out of millions of parameters). In particular, $S$ can forge malicious parameters that force the model to produce non-zero gradients for $\xi$ only when a specific adversarially-chosen property is present in the input batch used to compute the update. Then, the server can preserve the target's gradient for $\xi$ in the final aggregated value by forcing the non-target users to unconditionally produce zero gradients only for the parameters $\xi$. This allows $S$ to recover the target's gradient for $\xi$ in \TT{plaintext} and ascertain the presence of the queried property in the user's private data (\eg membership inference).
Eventually, this demonstrates that a malicious server can cast extremely effective property inference attacks on individual users under SA, while ensuring the stealthiness of the attack.

% We propose a second attack, dubbed \textbf{canary-gradient}.
% Here, a malicious server forges malicious parameters $\widetilde{\Theta}$ that force a user $u_i$ to produce a gradient $\Delta^{\widetilde{\Theta}}_{\dataset_i}$ that is $[0]$ only in a few adversarially chosen locations (in practice only the gradient of a subset of parameters $\xi$ of $\Theta$ will be set to $[0]$).
% As a consequence, the canary-gradient results to be stealthier than the gradient suppression attack.
% A gradient containing $[0]$ in a few locations is typical even in honest executions of FL.
% Naturally, being stealthy reduces the amount of information that a malicious server can recover.
% In particular, the canary-gradient attack ``only'' permits to execute property inference attacks.
% As for gradient suppression, a malicious server can use both canary-gradient and model inconsistency to bypass SA and target a specific user $u_\labeltarget$.
% In~\Cref{sec:attack2} we provide a detailed presentation.

\paragraph{The (in)correct usage of SA in FL}
In cryptographic terms, SA protocols are specialized multi-party computation (MPC) protocols that implement the ideal functionality $f^{\labelSecAggr}(v_1, \ldots, v_n) = \sum_{u_i \in \cUsers} v_i = v$. They are built assuming that the inputs of honest users are untamperable, i.e., an adversary has no control over the input $v_i$ of an honest user $u_i\in\cUsers$.
This holds in both the semi-honest and malicious security models.

Unfortunately, the above assumption does not hold in FL.
As discussed earlier, a malicious server $S$ can tamper with the inputs $(v_1, \ldots, v_n)$ of the honest users $\cUsers$.
Unlike other attacks, ours is the first that does not contradict the security of the underlying SA protocol and does not require auxiliary information about user data.
Our attacks will succeed regardless of the number of FL protocol users, which is additional evidence that SA and FL cannot protect against adversarial servers.

% Even assuming a perfectly secure SA, our attacks will succeed independently of the number of users running the FL protocol, showing that the combination of SA and FL is faulty and ineffective against an active adversarial server.

\paragraph{Preventing model inconsistency.}
%The FL vulnerability can be described as the following: Users implicitly assume that the parameters sent by the server are identical and will stay the same for different devices.
%This allows the server to execute model inconsistency and bypass the SA phase of FL.
We discuss ways to help mitigate model inconsistency by integrating consistency checks during the FL protocol.
In particular, we show that the two most influential SA protocols of Bonawitz et al.~\cite{bonawitz2017practical} (CCS'17) and Bell et al.~\cite{bell2020secure} (CCS'20) can be modified to incorporate consistency checks without affecting the efficiency of the original FL protocol.
This is achieved by linking SA's masking values (generated by a particular user to hide its input) to the parameters received from the server.
By doing it this way, when two or more users receive different parameters, their masks will look random. As a result, the malicious server cannot tell whose input is whose, reducing the attack efficacy.

Finally, we will discuss how DP techniques can be used in conjunction with SA, which remains a suitable solution to prevent a malicious server from breaching users' privacy.
We refer the reader to~\Cref{sec:def} for more details.
%\\

To make our results reproducible, we made our code available.\footnote{\url{https://github.com/pasquini-dario/EludingSecureAggregation}.}

\section{Related Work}
\label{sec:related-work}

% FL~\cite{shokri2015privacy,mcmahan2017communication} allows a server $S$ to train a deep neural network $f_{\Theta}$ on a dataset that is distributed among different users.
% At a high level, each user $u_i$ is required to locally train $f_{\Theta}$ using their local data $\dataset_i$ and send a model update $\Delta^{\Theta}_{\dataset_i}$ to $S$. The latter aggregates the users' updates $\{\Delta^{\Theta}_{\dataset_i}\}$ and computes the new parameters $\Theta'$ of $f$, accordingly.
% Multiple rounds of this process are executed until the parameters $\Theta$ converge to a final value.
% %To keep the training efficient,
% In its main form, FL leverages a star topology in which users only communicate with a central server $S$ (i.e., users do not have a communication channel between them) that synchronizes the parties and coordinates the execution of the protocol.
% FL has been proposed as a candidate solution to train deep neural networks while, at the same time, protecting users' datasets.
% Each user $u_i$ does not need to outsource their local dataset $\dataset_i$ during the training process.
% Indeed, $u_i$ only needs to to send, in each round, a model update $\Delta^{\Theta}_{\dataset_i}$ (e.g., gradients) to the server $S$.

\subsection{Federated learning and secure aggregation}\label{sec:related-work-SA}

The distributed architecture of FL protocol~\cite{shokri2015privacy,mcmahan2017communication} provides a fertile ground for attackers~\cite{hitaj2017deep,melis2019exploiting, 8835245, localpos}.
This is because a malicious party, mainly the server, has access to sensitive information such as model updates that can be exploited to violate users' privacy.
Accordingly, SA has been proposed by Bonawitz et al.~\cite{bonawitz2017practical} as a fundamental step to increase the security of FL without modifying the original structure of the protocol.
\iffull
In the setting of FL, SA is a specialized MPC protocol that allows the server to securely compute the sum of some values (known as aggregation) held by the users (e.g., integers, vectors).
The security definition of SA is the same as MPC: I.e., nothing is leaked except what can be inferred from the output (the sum of the values).
The deployment of SA in FL is beneficial since it allows us to compute the aggregation of users' model updates $\{\Delta^{\Theta}_{\dataset_i}\}$ (e.g., the gradients) without revealing to the server the individual updates of the users.
Keeping the individual updates private is essential since they carry sensitive information about the local training datasets of the users (Section~\ref{sec:gradient-inversion-fl}).
\fi
Subsequent works focus on the development of new SA protocols for FL with reduced communication/computation overhead~\cite{bell2020secure,choi2020communication,guo2020v,kadhe2020fastsecagg,so2021turbo}, multiple servers~\cite{beguier2020safer}, increased robustness against malicious updates~\cite{pillutla2019robust,burkhalter2021rofl}, or with verifiable aggregation~\cite{xu2019verifynet,guo2020v}.\footnote{In FL with verifiable aggregations~\cite{xu2019verifynet,guo2020v}, users are supposed to verify the integrity of the aggregated model updates and compute the new parameters (if the parameters are updated solely by the server, then the verifiability of aggregated gradients becomes meaningless). This differs from the original FL protocol~\cite{shokri2015privacy,mcmahan2017communication} in which the server updates the model for improved scalability.}

\ifconference
Separately, several other works focused on building new protocols (or propose significant modifications of FL, e.g., protocol, architecture, etc.) to train a deep neural network without leaking unnecessary information about the datasets of the users.
We refer the reader to~\Cref{sec:related-work-alternative-defenses} for more details.
\fi

\subsection{Gradient Inversion}\label{sec:gradient-inversion-fl}
A core privacy concern in FL is the role of the server.
In this direction, it has been shown that, without SA, even a semi-honest server can invert users' gradients (sent as a model update in a particular round of FL) and compute a close-enough approximation of users' local training datasets.
In a nutshell, by leveraging $f_{\Theta}$ (where $\Theta$ are the parameters of the current round of FL) and the gradient locally computed by the user $u$ using a subset $\dataset$ of its local data, the server can recover $\dataset$ by searching a set of instances $\widehat{\dataset}$ that generates the gradient similar to the one sent by the user.
Thanks to the inherent smoothness of the neural network $f_\Theta$, this searching problem can be defined as a second-order optimization, i.e.,
\begin{equation}
 \text{argmin}_{\widehat{\dataset}}[d(\nabla^\Theta_{\widehat{\dataset}}, \nabla^\Theta_{\dataset}) \cdot \alpha r(\widehat{\dataset})]
 \label{eq:opt}
\end{equation}
where $\widehat{\dataset}$ is the candidate solution of the malicious server~$S$, $d$~is a distance function to measure the discrepancy between the gradient signals $\nabla^\Theta_{\widehat{\dataset}}$ and $\nabla^\Theta_{\dataset}$, $r$ is a regularizer defined on the input domain, and $\alpha$ is the weight associated to the regularization term in the optimization.
In the work of Zhu~\etal~\cite{invg1}, $d$ is set to be the Euclidean distance, and the {L-BFGS} solver is used to solve the optimization problem.
The follow-up work of Geiping~\etal~\cite{invg2} improves their approach/results by noting that the gradient signal is scale-invariant and accounts for that in defining the optimization objective in~\Cref{eq:opt}.
\iffull
For this reason, they use the cosine similarity as distance function $d$ as well as a gradient-descend-based algorithm (\ie~Adam~\cite{adam}) to carry out the second-order optimization while regularizing $\widehat \dataset$ with total variation.\footnote{Their work focuses on the image domain.}
These modifications significantly improve
\fi
\ifconference
Their work improves
\fi
 the effectiveness of the inversion attack, drastically increasing its applicability on real-world architectures such as ResNets~\cite{resnet} and more realistic batch sizes and a number of FedAVG local iterations. These results have been further improved in the work of Yin~\etal~\cite{nvidiagi} by relying on additional regularization terms and tailored optimization techniques.

A more recent line of research dispensed with optimization-based approaches to focus on closed-form procedures to recover data from gradients in deep neural networks~\cite{rgap, pan2020theoryoriented}. In this vein, a recent work of Fowl~\etal~\cite{robbing} (which is concurrent to our work) improves over previous approaches by considering a malicious server that modifies the FL architecture and crafts the network parameters to artificially create a neural layer that retains information on the input batch.
In particular, this is a linear layer followed by a \relu~activation whose parameters must be chosen considering the private sets' CDF for a given property, i.e., the attacker must have some auxiliary information on users' training datasets.
Unfortunately, this extra knowledge may not be acquired in realistic scenarios (when users' data distributions are unknown) and weakens the applicability of the attack. In addition, the server needs to manipulate the model architecture and place a linear layer at the start of the network to maximize the attack effectiveness. However, this modification is unworkable for typical deep learning applications (\eg in computer vision).
% In this case, it can allow the server to directly reconstruct a verbatim copy of some of the user's training instances from gradient updates.

Lam~\etal~\cite{gdis} showed that, if SA is enabled, a malicious server $S$ can still try to reconstruct individual contributions by observing multiple training rounds of FL.
This disaggregation process can be reduced to a matrix factorization problem.
However, their attack is effective only in a particular restricted setting in which the malicious server $S$ $(i)$ alters the protocol execution by providing \emph{always} the same parameters $\widetilde{\Theta}$ at each round of FL (i.e., users compute their gradient updates always on the same model $f_{\widetilde{\Theta}}$), $(ii)$ leverages additional side-channel information about users' participation in the training rounds, and $(iii)$ users are required to use the same local training dataset at each round of FL.
%These two last conditions are particularly prohibitive and make the attack ineffective in the real world.
Their approach enables the gradient inversion attack of~\cite{invg1, invg2, nvidiagi} to scale and be effective in more realistic scenarios where there is a significant number of users participating in the protocol execution of FL.
Nevertheless, its feasibility still depends on the number of active users, network parameters, and other factors such as the number of rounds that the server monitors to recover individual gradients accurately.
Additionally, while the authors showed that this approach could handle noise, its applicability is inherently limited in FedSGD, where local training is performed on randomly selected batches rather than the entire, static, local dataset. Finally, this attack is not applicable in large-scale, real-world deployments of FL~\cite{gboard0, gboard1}, where users participate in the protocol only once with a high probability.
\iffull
\subsection{Other protocols}\label{sec:related-work-alternative-defenses}
\iffull
Several other works focused on building new protocols (or propose major modifications of FL, e.g., protocol, architecture, etc.) to \emph{train a deep neural network without leaking unnecessary information about the datasets of the users}.
\fi
% Other works focus on proposing new protocols or modifications for FL by leveraging
% leverage other cryptographic primitives (such as homomorphic encryption, functional encryption, multi-party computation, or fully homomorphic encryption) to preserve the secrecy of users updates.\footnote{In some works (such as~\cite{xu2019hybridalpha,truex2019hybrid}), these primitives are used to implement a secure aggregation of users' model updates.}
Aono et al.~\cite{aono2017privacy} leverage a homomorphic encryption scheme to permit users to share encrypted model updates (e.g., gradients) with the server. The latter will compute the new model parameters in an encrypted fashion, using the additive homomorphic properties of the underlying encryption scheme. The new parameters remain encrypted on the server-side.

Truex et al.~\cite{truex2019hybrid} use a similar approach by leveraging the homomorphic properties of Pailler's cryptosystem. Here, the Pailler encryption scheme is necessary to allow the server to aggregate the model updates in an encrypted form. In contrast to~\cite{aono2017privacy}, the aggregation will be decrypted by the users and made public to the server.

An analogous approach is used by EastFly~\cite{dong2020eastfly} and BatchCrypt~\cite{zhang2020batchcrypt} with the main difference of keeping the aggregated updates secret to the server: The aggregation is locally decrypted by the users that will also compute the new model parameters (this differs from the original FL protocol~\cite{shokri2015privacy,mcmahan2017communication} in which the parameters are stored and updated by the server).
In addition,~\cite{zhang2020batchcrypt} proposes a batching encoding scheme (that preserves the homomorphic properties of the underlying homomorphic encryption schemes) to reduce the number of encryption operations and speed up the efficiency of the protocol.

HybridAlpha~\cite{xu2019hybridalpha} uses a multi-input functional encryption scheme to compute the aggregation.
Informally, each user $u_i$ computes the encryption $Enc_{pk_i}(\Delta^{\Theta}_{\dataset_i})$ of its update $\Delta^{\Theta}_{\dataset_i}$. Once the server has received all the ciphertexts, it computes the aggregation by decrypting them using the decryption key $sk_f$ for the functionality $f(\Delta^{\Theta}_{\dataset_1}, \ldots, \Delta^{\Theta}_{\dataset_n}) = \sum^{n}_{i=1} \Delta^{\Theta}_{\dataset_i}$.
Analogously, SAFElearn~\cite{fereidooni2021safelearn} leverages either multi-party computation or fully homomorphic encryption to protect the individual's updates of users.

Poseidon~\cite{sav2020poseidon} significantly deviates from the original FL architecture.
In such a system, users are organized according to a tree hierarchy, and a multi-key fully homomorphic encryption scheme is used to protect users' updates and the parameters of the neural network.
At each round of the training phase, the root user sends the encrypted parameters to all users that, in turn, compute their updates according to their local training data.
Then, each encrypted user's update is sent to the parent that will then aggregate all children's updates. At the end of this recursive process, the root receives the encrypted aggregation (that contains the updates of all users) and will update the parameters.
The entire computation of the training process is executed inside the multi-key fully homomorphic encryption scheme.
This permits them to keep the updates and the parameters encrypted.
The latter remains encrypted even after the training process, and the model evaluation requires further computation inside the multi-key fully homomorphic encryption scheme.
Note that Poseidon~\cite{sav2020poseidon} is an extension of SPINDLE~\cite{froelicher2021scalable}.
The former handles complex machine learning models (such as neural networks), while the latter supports only generalized linear models.

Lastly, we mention Cerebro~\cite{zheng2021cerebro} that proposes a compiler to automatically transform Python-like domain-specific language into an optimized MPC protocol for collaborative learning allowing users to keep their plaintext data secret. We stress that Cerebro~\cite{zheng2021cerebro} does not relate in any way with FL~\cite{shokri2015privacy,mcmahan2017communication} and it must be interpreted as an alternative to FL.
\paragraph{Attack applicability}
As discussed in Section~\ref{sec:threat-model}, our attacks are general and equally effective independently from the SA protocol used. 
This because they exploit a vulnerability of FL caused by the incorrect usage of SA.
	% target the ideal functionality of secure aggregation, and, therefore, they apply to every SA protocol.
	More precisely, our main attack applies to all the FL/SA protocols that do not prevent the parameter server from deviating from the honest execution. This class also includes most of the schemes that rely on fully homomorphic encryption (FHE) \eg~\cite{fereidooni2021safelearn}. The reason is that FHE would still allow the server to multiply the individual encrypted parameters by the constant $0$ and produce $0$ gradient everywhere when used by the non-target users (see Appendix~\ref{sec:gsea}). The only requirement is that server is required to collude with a least one user in order to access the result of the attack (the target model update) once decrypted by the pool (\eg $m>0$).

Protocols based on Trusted Execution Environment (TEE) can stop the server from executing malicious code~\cite{PPFL, hashemi2021byzantine}. Thus, a properly deployed TEE environment with ideal/perfect hardware and secure and authenticated communication would prevent all the active attacks currently in the literature~\cite{robbing, gdis, papernot}, including ours. However, the reality is that trusted hardware is vulnerable to side-channel attacks, and there is significant performance degradation when extending side-channel protections to arbitrary computations. Therefore, it remains unclear whether side-channel attacks can be entirely eliminated.

\fi

\section{Preliminaries}\label{sec:preliminaries}
\iffull
\subsection{Notation}\label{sec:notation}
\fi
We use small letters (such as $x$) to denote concrete values, calligraphic letters (such as $\mathcal{X}$) to denote sets.
For a string $x \in \bin^*$, we let $|x|$ be its length; if $\mathcal{X}$ is a set, $|\mathcal{X}|$ represents the cardinality of $\mathcal{X}$.
In the setting of deep learning, we use the notation $[\cdot]$ to express a tensor (i.e., vector) of arbitrary dimension.
We write $[x]$ for a tensor filled with the value $x$.
When a set is included between square brackets (\eg $[\mathbb{R}^+]$), the tensor is filled with arbitrary elements from that set.
\iffull
We model cryptographic algorithms (e.g., adversary) as (possibly interactive) Turing machines.
If $A$ is a deterministic (resp. randomized) algorithm, we write $y = A(x)$ \iffalse (resp. $y\getsr A(x)$) \fi to denote a run of $A$ on input $x$ and output $y$; if $A$ is randomized, $y$ is a random variable.
An algorithm $A$ is probabilistic polynomial-time (PPT) if $A$ is randomized and for any input $x \in \bin^*$ the computation of $A(x)$ terminates in a polynomial number of steps (in the input size).
We denote by $\lambda \in \NN$ the security parameter of cryptographic primitives, and we implicitly assume that every algorithm takes as input the security parameter (written in unary).
A function $\nu : \NN \rightarrow [0, 1]$ is called negligible in the security parameter $\lambda$ if it vanishes faster than the inverse of any polynomial in $\lambda$, i.e. $\nu(\lambda) \in O(1/p(\lambda))$ for all positive polynomials $p(\lambda)$.
We write $\negl$ to denote an unspecified negligible function in the security parameter. Similarly, we write $\poly$ to denote all possible polynomials $p(\lambda)$.
Let $X$ and $Y$ be two random variables.
We say that $X$ and $Y$ are computationally indistinguishable, denoted $X\approx_c Y$, if for all PPT distinguishers $D$ we have
\[
  \left\vert \prob{D(1^\secpar, X) = 1} - \prob{D(1^\secpar, Y) = 1} \right \vert \leq \negl.
\]

\fi
\subsection{Neural networks}\label{sec:nn}
\iffull
A neural network is a differentiable non-linear mapping $f_{\Theta} : \cX \rightarrow \cY$ where $\Theta \in \RR^m$ are the parameters defining the mapping $f_{\Theta}$, $\cX$ is the input space, and $\cY$ is the output space (e.g., class labels). The pool of representable functions is defined by the so-called \emph{architecture} of the network that is generally specified as a sequence of logic partitions called \emph{layers} \ie non-linear parametric functions on their own. Within the paper, we abstract a neural layer with respect to some parameters $\Theta \in \RR^m$ using the following notation:
\begin{equation}\label{eq:single_layer}
	\ell(x) = \phi(x \otimes \theta + b),
\end{equation}
where the symbols $\theta \in \Theta$ and $b \in \Theta$ are arbitrarily shaped real tensors\footnote{The shape of these tensors depend on the operator $\otimes$.} that represent the learnable parameters of the layer $\ell$. Hereafter, we refer to $\theta$ and $b$ as the \emph{kernel} and the \emph{bias}, respectively.
The operation $\otimes$ abstracts the application of the kernel on the input tensor $x$.
As an example, $\otimes$ can be a matrix multiplication operator ($\ell$ is a fully connected layer), a convolution operator ($\ell$ is a convolutional layer), or a more complex parametric transformation such as the multi-head attention mechanism used to build transformer networks~\cite{attention}.
This definition captures other core building blocks such as normalization layers. The function~$\phi$, instead, is the activation function of $\ell$, which makes $\ell$ non-linear.

A deep neural network, in turn, is a function $f_{\Theta} : \cX \rightarrow \cY$ defined by the composition of many layers, i.e.,
\[
f_\Theta(x)=\ell_{n-1}(\dots(\ell_1(\ell_0(x))))
\]
where $\ell_{i}(x) = \phi(x \otimes \theta_i + b_i)$ (as defined in~\Cref{eq:single_layer}), $\theta_i \in \Theta$, $b_i \in \Theta$, and $x \in \cX$.
Deep neural networks are powerful function approximators capable of accurately describing relations among high-dimensional spaces.
Once a target function is chosen, a differentiable loss function $\loss$ is defined and used to guide the approximation.
This process, known as \emph{learning}, consists of finding the configuration of parameters $\Theta$ that minimizes the discrepancy between the target function and the neural network $f_\Theta$ by relying on gradient-descent-based optimization techniques such as Stochastic Gradient Descent (SGD).
In particular, for each layer $\ell_i$, we compute the partial first-order derivative of the loss function (which is a function of $\ell_i$) with respect to $\theta_i$ and $b_i$, obtaining $\frac{\partial\loss}{\partial \theta}$ and $\frac{\partial\loss}{\partial b}$. Then, we use these information to perform gradient descent on $\theta_i$ and $b_i$, updating them to the new configurations $\theta'_i$ and $b'_i$, respectively.
This also yields a new configuration $\Theta'$ of the global parameters of the deep neural network $f$.

The (unknown) target function we want to approximate is usually described by a dataset $\dataset$ of the form $\dataset = \{(x_i, y_i)\}$ where $x_i \in \cX$ is an input and $y_i \in \cY$ is the output (i.e., classification) of the target function on input $x_i$.
In such a case, the learning goes through as described above, where the loss function $\loss$ simply measures the discrepancy between $f_\Theta(x_i) = h_i$ and $y_i$ where $\Theta$ are the parameters of the current round of the learning process.

In the rest of the paper, we use the terms ``example'' or ``instance" interchangeably to denote an input $x_i$ of $\dataset$.
Also, we make the following abuse of notation $x_i \in \dataset$ to say that $x_i$ is an instance contained in the dataset $\dataset$ (i.e., we ignore labels).
\fi

\ifconference
% \paragraph{\new{Neural Networks}}
We abstract a neural layer with respect to some parameters $\Theta \in \RR^m$ using the following notation:
\begin{equation}\label{eq:single_layer}
	\ell(x) = \phi(x \otimes \theta + b),
\end{equation}
where the symbols $\theta \in \Theta$ and $b \in \Theta$ are arbitrarily shaped real tensors\footnote{The shape of these tensors depend on the operator $\otimes$.} that represent the learnable parameters of the layer $\ell$. Hereafter, we refer to $\theta$ and $b$ as the \emph{kernel} and the \emph{bias}, respectively.
The operation $\otimes$ abstracts the application of the kernel on the input tensor~$x$.
As an example, $\otimes$ can be a matrix multiplication operator ($\ell$ is a fully connected layer), a convolution operator ($\ell$ is a convolutional layer), or a more complex parametric transformation such as the multi-head attention mechanism used to build transformer networks~\cite{attention}.
This definition captures other core building blocks such as normalization layers. The function~$\phi$, instead, is the activation function of $\ell$, which makes $\ell$ non-linear.

A deep neural network is a function $f_{\Theta} : \cX \rightarrow \cY$ defined by the composition of many layers, i.e.,
$f_\Theta(x)\myeq \ell_{n-1}(\dots(\ell_1(\ell_0(x))))$ where $\ell_{i}(x) \myeq \phi(x \otimes \theta_i + b_i)$ (as defined in~\Cref{eq:single_layer}), $\theta_i \in \Theta$, $b_i \in \Theta$, and $x \in \cX$.
\fi

\subsection{Federated learning}\label{sec:federate-learning}
Federated learning (FL) allows a set of users $\cUsers~=~\{u_1, \ldots, u_n\}$, where each $u_i$ holds a local training dataset $\dataset_i$, to train the deep neural network $f_\Theta: \cX \rightarrow \cY$ on a global dataset that is distributed among~$\cUsers$.
A centralized server $S$ coordinates the communications between the users to train a deep neural network on $\dataset = \bigcup_{u_i \in \cUsers} \dataset_i$ in such a way that each $\dataset_i$ does not leave the $u_i$'s device.
The learning phase of FL is an interactive process that is divided into rounds.
In the setting of FL, we denote with $t$ the current round of FL and we use the superscript ``$^{(t)}$'' (sometimes in combination with the subscript ``$_i$'') to denote values used or generated (such as batches, model updates, model parameters) during the current round $t$ (by a particular user $u_i$).
At each round $t\in \NN$, the server $S$ (holding the parameters $\Theta^{(t)}$ of the deep neural network $f$) 
sends to the subset of available users $\cUsers^{(t)} \subseteq \cUsers$ the parameters $\Theta^{(t)} \in \RR^m$.
%selects a subset of users $\cUsers^{(t)} \subseteq \cUsers$ and it sends them the parameters $\Theta^{(t)} \in \RR^m$.

The set $\cUsers^{(t)}$ is composed of the users entitled to participate in the learning phase during the current round $t$.
Each $u_i$ samples a random subset $\dataset^{(t)}_{i}$ (known as batch) from its dataset $\dataset_i$ and locally trains $f_{\Theta^{(t)}}$ on $\dataset_i$.
The final result of the local training is a model update $\Delta^{\Theta^{(t)}}_{\dataset^{(t)}_i})$ that is then forwarded to the server $S$.
The latter will be responsible of computing the new configuration $\Theta^{(t+1)}$ of the parameters with respect to average of the model updates $\{\Delta^{\Theta^{(t)}}_{\dataset^{(t)}_i}\}_{u_i \in \cUsers}$ received by the server $S$.
This process is iterated until the parameters $\Theta$ converge.

\paragraph{FedSGD and FedAVG}
The computation of the users' model updates $\Delta^{\Theta^{(t)}}_{\dataset^{(t)}_i}$ and model parameters $\Theta^{(t)}$ vary according to the type of FL that is in place.
The two main approaches are known as \emph{federated stochastic gradient descent (FedSGD)} and \emph{federated averaging (FedAVG)}.
In FedSGD, a single step of gradient descend is performed per round $t \in \NN$.
In other words, the model update $\Delta^{\Theta^{(t)}}_{\dataset^{(t)}_i}$ (computed by a user $u_i$) corresponds to the gradients $\nabla^{\Theta^{(t)}}_{\dataset^{(t)}_i}$ computed with respect to the randomly chosen batch $\dataset^{(t)}_i \subseteq \dataset_i$.
This gradient $\nabla^{\Theta^{(t)}}_{\dataset^{(t)}_i}$ is set to be the model update $\Delta^{\Theta^{(t)}}_{\dataset^{(t)}_i}$ of user $u_i$.
On the server side, the single step of gradient descend is executed in order to compute the new model parameter $\Theta^{(t+1)}$ from $\Theta^{(t)}$ and $\{\Delta^{\Theta^{(t)}}_{\dataset^{(t)}_i}\}_{u_i \in \cUsers}$.
More formally, the parameters are updated as follows:
\iffull
\begin{align}\label{eq:model-update-SGD}
\Theta^{(t+1)} &= \Theta^{(t)} - \eta \frac{\sum_{u_i \in \cUsers^{(t)}} \Delta^{\Theta^{(t)}}_{\dataset^{(t)}_i}}{|\cUsers^{(t)}|} = \Theta^{(t)} - \eta \frac{\sum_{u_i \in \cUsers^{(t)}} \nabla^{\Theta^{(t)}}_{\dataset^{(t)}_i}}{|\cUsers^{(t)}|},
\end{align}
where $\eta$ is the learning parameter.
\fi
\ifconference
\begin{align}\label{eq:model-update-SGD}
\Theta^{(t+1)} &= \Theta^{(t)} - \eta \frac{\sum_{u_i \in \cUsers^{(t)}} \Delta^{\Theta^{(t)}}_{\dataset^{(t)}_i}}{|\cUsers^{(t)}|}
\end{align}
where $\Delta^{\Theta^{(t)}}_{\dataset^{(t)}_i} = \nabla^{\Theta^{(t)}}_{\dataset^{(t)}_i}$ and $\eta$ is the learning parameter.
\fi

On the other hand, in FedAVG, users locally perform $k \in \NN$ iterations of stochastic gradient descent, producing new model parameters at each round.
More formally, let $\Theta^{(t)} = \Theta^{(t,1)}_i$ be the parameters received by $u_i$ from the server at the beginning of round $t$.
For every $j \in \{1,\ldots,k\}$, a user $u_i$ samples a random batch $\dataset^{(t,j)}_i\subset \dataset_i$ and computes a gradient $\nabla^{\Theta^{(t,j)}}_{\dataset^{(t,j)}_i}$.
Then, it locally updates the model parameters as defined in~\Cref{eq:model-update-SGD}, i.e.,
$\Theta^{(t,j+1)}_i = \Theta^{(t,j)}_i - \eta \cdot \nabla^{\Theta^{(t,j)}}_{\dataset^{(t,j)}_i}$.
Then, after $k$ iterations of gradient descent, the final model parameters $\Theta^{(t,k)}_i$ will be the model update $\Delta^{\Theta^{(t)}}_{\dataset^{(t)}_i}$ that user $u_i$ will send to the server.
Finally, the server $S$ only needs to compute the new parameters $\Theta^{(t+1)}$ that is the average of the parameters received from $\cUsers^{(t)}$:
\begin{equation}\label{eq:model-update-AVG-server}
\Theta^{(t+1)} = \frac{\sum_{u_i \in \cUsers^{(t)}} \Delta^{\Theta^{(t)}}_{\dataset^{(t)}_i}}{\sum_{u_i \in \cUsers^{(t)}} b_i} = \frac{\sum_{u_i \in \cUsers^{(t)}} \Theta^{(t,k)}_i}{\sum_{u_i \in \cUsers^{(t)}} b_i},
\end{equation}
where $b_i = \sum^k_{j = 1} |\dataset^{(t,j)}_i|$.\footnote{Observe that, in order to compute the new model parameters $\Theta^{(t+1)}$, the server needs to receive either $b_i = \sum^k_{j =1} |\dataset^{(t,j)}_i|$ from each user $u_i$ or the size of each $\dataset^{(t,j)}_i$ needs to be fixed.}
\iffull
The complete description of the FL protocol (with FedSGD or FedAVG) is depicted in~\Cref{fig:FL}.
\begin{figure*}
  \small
\begin{protocolenv}{Federated Learning}
\textit{Inputs.} For every $u_i\in\cUsers$, user $u_i$ holds a local training dataset $\dataset_i = \{(x_j, y_j)\}_{j\in\{1,\ldots,\ell\}}$.
The server $S$ fixes the architecture $f$ of the deep neural network, the learning parameter $\eta \in [0,1]$, and the initial parameters $\Theta^{(1)}$.
\sbline
\textit{Goal.} The server $S$ obtains the final model parameters $\Theta$ of the deep neural network $f$.
\sbline
\textit{The protocol:}
\begin{itemize}
 \item \textbf{For $t \in \NN$, until $\Theta^{(t)}$ converges.}
 \begin{enumerate}
  \item $S$ samples a random subset of users $\cUsers^{(t)} \subseteq \cUsers$ that will participate in training of $f$ in the current round $t$.
  $S$ sends the parameters $\Theta^{(t)}$ to each $u_i \in \cUsers^{(t)}$.

  \item Each user $u_i \in \cUsers^{(t)}$ receives $\Theta^{(t)}$ and proceeds as follows:
  \begin{enumerate}
    \item \textbf{If FedSGD.} $u_i$ samples a random training batch $\dataset^{(t)}_i \subseteq \dataset_i$ and computes the gradient $\nabla^{\Theta^{(t)}}_{\dataset^{(t)}_i}$. Finally, it sets its model update $\Delta^{\Theta^{(t)}}_{\dataset^{(t)}_i} = \nabla^{\Theta^{(t)}}_{\dataset^{(t)}_i}$.
    \item \textbf{If FedAVG.} Let $\Theta^{(t,1)}_i = \Theta^{(t)}$ For $j \in \{1,\ldots,k\}$, $u_i$ samples a random training batch $\dataset^{(t,j)}_i \subseteq \dataset_i$ and computes the gradient $\nabla^{\Theta^{(t,j)}}_{\dataset^{(t,j)}_i}$. Then, it updates the model parameters by computing $\Theta^{(t,j+1)}_i = \Theta^{(t,j)}_i - \eta \cdot \nabla^{\Theta^{(t,j)}}_{\dataset^{(t,j)}_i}$.
    Finally, it sets $\Delta^{\Theta^{(t)}}_{\dataset^{(t)}_i} = \Theta^{(t,k)}_i$ and $b_i = \sum^k_{j =1} |\dataset^{(t,j)}_i|$.
  \end{enumerate}

  \item Each user $u_i \in \cUsers^{(t)}$ sends its model update as follows:
  \begin{enumerate}
    \item \textbf{If SA disabled.} It sends $\Delta^{\Theta^{(t)}}_{\dataset^{(t)}_i}$ to $S$. In addition, \textbf {if FedAVG}, $u_i$ sends $b_i$ to $S$.
    \item \textbf{If SA enabled.} It provides the input $\Delta^{\Theta^{(t)}}_{\dataset^{(t)}_i}$ to $f^{\labelSecAggr}$. In addition, \textbf {if FedAVG}, $u_i$ sends $b_i$ to $S$.
  \end{enumerate}

  \item $S$ computes the new model parameters $\Theta^{(t+1)}$ in the following way:
  \begin{enumerate}
    \item \textbf{If SA disabled.}
    \begin{enumerate}
      \item \textbf{If FedSGD.} $S$ receives $\{\Delta^{\Theta^{(t)}}_{\dataset^{(t)}_i}\}_{u_i \in \cUsers}$ from users and computes
      $\Theta^{(t+1)} = \Theta^{(t)} - \eta \cdot v/b$
      where $v = \sum_{u_i \in \cUsers^{(t)}} \Delta^{\Theta^{(t)}}_{\dataset^{(t)}_i}$ and $b = |\cUsers^{(t)}|$.\label{itm:step-sgd}
      \item \textbf{If FedAVG.} $S$ receives $\{(\Delta^{\Theta^{(t)}}_{\dataset^{(t)}_i},b_i)\}_{u_i \in \cUsers}$ from users and computes
      $\Theta^{(t+1)} = v/b$ where $v = \sum_{u_i \in \cUsers^{(t)}} \Delta^{\Theta^{(t)}}_{\dataset^{(t)}_i}$ and $b = \sum_{u_i \in \cUsers^{(t)}} b_i = \sum_{u_i \in \cUsers^{(t)}} \sum^{k}_{j =1} |\dataset^{(t,j)}_i|$. \label{itm:step-avg}
    \end{enumerate}
    \item \textbf{If SA enabled.} $S$ receives $v = \sum_{u_i \in \cUsers^{(t)}} \Delta^{\Theta^{(t)}}_{\dataset^{(t)}_i}$ from $f^\labelSecAggr$. Then, it proceeds as follows:
    \begin{enumerate}
      \item \textbf{If FedSGD.} $S$ computes $\Theta^{(t+1)}$ as in~\Cref{itm:step-sgd} using $v$ outputted by $f^\labelSecAggr$.
      \item \textbf{If FedAVG.} $S$ receives $\{b_i\}_{u_i \in \cUsers}$ from users and computes $\Theta^{(t+1)}$ as in~\Cref{itm:step-avg} using $v$ outputted by $f^\labelSecAggr$.
    \end{enumerate}
  \end{enumerate}
 \end{enumerate}
\end{itemize}
\end{protocolenv}
\caption{Description of FedSGD-based and FedAVG-based FL with SA either enabled or disabled. The execution of SA is represented by the ideal functionality $f^\labelSecAggr$.}
\label{fig:FL}
\end{figure*}

\fi

\ifconference
\noindent
The complete description of the FL protocol (with FedSGD or FedAVG) is depicted in~\Cref{fig:FL} of the appendix.
\fi

\subsection{Secure aggregation}\label{sec:secure-aggregation}
A secure aggregation (SA) protocol is a specialized multi-party computation (MPC) protocol that allows a set of users to compute the summation (a.k.a. aggregation) of their inputs.
Let $\cUsers = \{u_1, \ldots, u_n\}$ be a set of users, each holding a secret input $v_i$ (e.g., integer, group element, vector).
A protocol $\Pi$ is a secure SA protocol if it securely implements the following ideal functionality:
\begin{equation}\label{eq:ideal-func-sa}
 f^{\labelSecAggr}(v_{1}, \ldots, v_{n}) = (v, \ldots, v)\quad \text{for } v = \sum_{u_i \in \cUsers} v_{i},
\end{equation}
i.e., a trusted third party executes $f^{\labelSecAggr}$ computes and returns to all users $\cUsers$ the aggregation $v$ of the users' inputs $(v_1, \ldots, v_n)$.
\iffull
The security of SA follows the standard ideal and real-world paradigm of MPC~\cite{cramer2005multiparty}.
Below, we present the ideal and real paradigm for a generic protocol implementing a functionality $f(x_1,\ldots,x_n) = (y_1,\ldots,y_n)$ where $y_i$ is the output returned to user $u_i$.

\ifconference
\paragraph*{Additional notation}

We now describe the ideal and real-world paradigm of MPC~\cite{cramer2005multiparty} that defines the security of MPC protocols (including SA).
\fi

\paragraph*{Real world} The real world refers to the scenario in which the real protocol $\pi$ is executed between the users $\cUsers$.
During its execution the parties $\cUsers$ exchange messages between themself in order to compute the $f(x_1, \ldots, x_n) = (y_1,\ldots,y_n)$.
In this setting, we assume the presence of an adversary $A$ that can be either semi-honest or malicious.
A semi-honest $A$ can take control of a subset of users $\widetilde{\cUsers} \subset \cUsers$.
This will allow $A$ to have access to their inputs $\{x_i\}_{u_i \in \widetilde{\cUsers}}$, the messages received, and the final output $f_i(x_1, \ldots, x_n) = y_i$.
In addition, if $A$ is malicious, then it can also program a corrupted user $u_i \in \widetilde{\cUsers}$ to deviate from the original protocol specification (e.g., send malicious messages).
We use the notation $V^\pi_i(x^*)$ to denote the view of the $i$-th party, i.e.,
\[
  V^\pi_i(x^*) = (x_i, r_i, m_1, \ldots, m_k)
\]
where $x^* = (x_1, \ldots, x_n)$, $r_i$ are the (private) random coins of $u_i$, and $m_j$ is the $j$-th message received by $u_i$.

\paragraph*{Ideal world} In the ideal world, the protocol execution is replaced by a trusted third party that honestly computes $f$.
In more detail, each user $u_i$ sends its input $x_i$ to the third party. The latter computes $f(x_1,\ldots, x_n) = (y_1,\ldots, y_n)$ and returns $y_i$ to each user $x_i$.
Even in this setting, the (either semi-honest or malicious) adversary $A$ can corrupt a set of parties $\widetilde{\cUsers}$ and, in turn, obtains their private inputs $\{x_i\}_{u_i\in\widetilde{\cUsers}}$ and outputs $\{y_i\}_{u_i\in\widetilde{\cUsers}}$.

\paragraph*{Security} At a high level, the security of $\pi$ is defined by comparing the ideal and real world. In particular, $\pi$ is secure if there exists a simulator $S$ that simulates the view of a real-world adversary $A$ by leveraging the interactions performed in the ideal world.
This implies that any attack in the real world ``corresponds'' to an attack in the ideal world that, in turn, provides the maximum security guarantees that can be achieved.
In other words, a secure protocol $\pi$ provides at least the same level of security as if we had an honest, trusted party computing $f$ without exposing to an adversary the inputs of uncorrupted users.

\begin{definition}
  Let $\pi$ and $\cUsers = \{u_1, \ldots, u_n\}$ be a $n$-party protocol that correctly computes an $n$-inputs $n$-outputs function $f$ and the set of users that participate in the protocol execution, respectively.
  Let $x_i$ be the input of user $u_i \in \cUsers$ and let $x^* = (x_1, \ldots,x_n)$.
  For $\widetilde{\cUsers} = \{u_{i_1}, \ldots, u_{i_k}\} \subset \cUsers$, we let $V^\pi_{\widetilde{\cUsers}}(x^*) = (\widetilde{\cUsers}, V^\pi_{i_1}(x^*), \ldots, V^\pi_{i_k}(x^*))$ and $f_{\widetilde{\cUsers}}(x^*) = (y_{i_1}, \ldots, y_{i_k})$.
  We say that $\pi$ securely computes $f$ if there exists a PPT simulator $S$ such that, for every $\widetilde{\cUsers} \subset \cUsers$, for every input $x^* = (x_1, \ldots, x_n)$, we have:
  \begin{align*}
  \{S(\widetilde{\cUsers}, \tilde{x}, f_{\widetilde{\cUsers}}(x^*))\}_{x^* \in \bin^*} \approx_c \{V^\pi_{\widetilde{\cUsers}}(x^*)\}_{x^* \in \bin^*}
  \end{align*}
  where $\tilde{x} = (x_{i_1}, \ldots, x_{i_k})$ for $u_{i_j} \in \widetilde{\cUsers}$.
\end{definition}

\noindent The security of $\Pi$ does not guarantee that nothing is leaked about other users' inputs.
Instead, it implies that nothing is leaked except what can be inferred from the final aggregation.
The information that can be inferred is highly correlated to the inputs provided to $f^\labelSecAggr$ (e.g., entropy).
\fi
\ifconference
Informally, a SA protocol $\Pi$ is considered secure if it is at least as secure as invoking the ideal functionality $f^\labelSecAggr$.
This is formalized using the standard ideal and real-world paradigm of MPC~\cite{cramer2005multiparty} (see~\Cref{sec:security-SA}).
The security of $\Pi$ does not guarantee that nothing is leaked about other users' inputs.
Instead, it implies that nothing is leaked except what can be inferred from the final aggregation.
The information that can be inferred is highly correlated to the inputs provided to $f^\labelSecAggr$ (e.g., entropy).
\fi

To increase the security of FL and mitigate attacks such as gradient inversion (see~\Cref{sec:related-work}), Bonawitz et al.~\cite{bonawitz2017practical} propose a communication-efficient, dropout resilient SA for FL.\footnote{In~\cite{bonawitz2017practical} the aggregation $v$ is obtained by the server $S$ only.
This can be represented as an ideal functionality $f^\labelSecAggr(v_1,\ldots, v_n, \bot) = (\bot,\ldots, \bot, v)$ where the $(n+1)$-th input/output is associated to the server $S$ and $\bot$ represents the empty string.}
When applied to FL, the protocol guarantees a secure aggregation of users' model updates -- sensitive values that carry important information about the users' datasets.
Hence, an adversary (e.g., malicious user or server) can not observe the individual model update $\Delta^{\Theta^{(t)}}_{\dataset^{(t)}_i}$ of a target user $u_i$.
This decreases the amount of information about the local training dataset $\dataset_i$ of the target user $u_i$ that an adversary can leak (e.g., by performing gradient inversion attacks).
We stress that SA protocols~\cite{bonawitz2017practical,bell2020secure} have the property of being robust: The final aggregation can always be computed when at most $\delta$ users drop out during the execution of the SA protocol (i.e., more than $n-\delta$ users are online).
When this condition is failed (i.e., the online users are less than $n-\delta$), the server cannot recover the final aggregated value. 
Hence, SA also guarantees that the server can only see the final aggregation when the latter contains at least $n-\delta$ model updates (ensuring the desired level of \TT{privacy by aggregation}).
\iffull
The overall FL protocol with SA enabled is depicted in Figure~\ref{fig:FL}.
\fi

\textbf{Both Bonawitz et al.~\cite{bonawitz2017practical} and Bell et al.~\cite{bell2020secure} (the most influential SA protocols in the literature) demonstrate the security of their protocols in both the semi-honest and malicious models} (including collusion between server and users)~\cite[Theorems 6.2 and 6.6]{bonawitz2017practical} and~\cite[Theorems 3.6 and 4.9]{bell2020secure}.
However, we emphasize that the results in~\cite{bonawitz2017practical,bell2020secure} cover only the security of the SA protocol.
Nothing is claimed about the security of the overall FL protocol with SA enabled.
%However, we emphasize that the results in~\cite{bonawitz2017practical,bell2020secure} cover only the security of the SA protocol, overlooking the security of the overall FL protocol with SA enabled.

\section{Threat model}\label{sec:threat-model}
In this section we formalize the threat model in which our attacks are defined (Section~\ref{sec:attack1} and Section~\ref{sec:attack2}).
{We adopt the \emph{exact} same threat model for which the SA protocols of Bonawitz et al.~\cite[Section 6.2]{bonawitz2017practical} (CCS '17) and Bell et al.~\cite[Section 4]{bell2020secure} (CCS '20) have been demonstrated to be secure: A malicious parameter server (aggregator) that can corrupt at most a fixed number $m$ (out of $n$) of users.}\footnote{The number of corrupted users $m$ depends on the implementation of the SA protocol. For instance, this is $\lceil \frac{n}{3} \rceil-1$ for  Bonawitz et al.~\cite{bonawitz2017practical}.}
Observe that these SA protocols~\cite{bonawitz2017practical,bell2020secure} are the most influential and practical-oriented solutions in the field.

More formally, in each round $t \in \NN$, the active users $\cUsers^{(t)}$ do not send their model updates $\{ \Delta^{\Theta^{(t)}}_{\dataset^{(t)}_i}\}_{u_i \in\cUsers^{(t)}}$ in the clear.
Instead, they execute the SA protocol $\Pi$ to securely compute the aggregation $v = \sum _{u_i \in \cUsers^{(t)}} \Delta^{\Theta^{(t)}}_{\dataset^{(t)}_i}$ (see~\Cref{sec:secure-aggregation}) for at least $q$ users. Only the final aggregation $v$ is revealed to the server~$S$.
More importantly, we do not target any specific implementation of SA. 
In fact, our attack exploits a vulnerability caused by the incorrect usage of SA in the FL protocol. In particular, the FL protocol does not validate the inputs of SA (i.e., model updates).
For this reason, to keep the attack general, we replace the execution of the underlying SA protocol with the invocation of its ideal functionality $f^\labelSecAggr$ (\Cref{sec:secure-aggregation}).
\par

We model an adversarial server $S$ whose objective is to learn information about the local dataset $\dataset_\labeltarget$ of one (or more) target user $u_\labeltarget$ that participates in the execution of the protocol, outside what can be learned from the aggregated model updates. $S$ can deviates from a honest execution (e.g., it sends arbitrary messages) as defined in~\cite[Section 6.2]{bonawitz2017practical} and~\cite[Section 4]{bell2020secure}.
Specifically, in our attacks, the malicious server $S$ exploits the model inconsistency attack vector; that is, $S$ provides arbitrary malicious parameters to arbitrary users even within the same training round $t \in \NN$.
Although the adversarial server $S$ is allowed to collude with at most $m$ users (as considered by~\cite{bonawitz2017practical,bell2020secure}), our attacks demonstrate that SA is ineffective in FL even when $m \myeq 0$, i.e., the server is malicious, but it does not collude with any user.\footnote{However, collusion with users improves the effectiveness of the attacks when additional mechanisms such as distributed differential privacy are in place.}
This ensures the effectiveness of the proposed attacks also outside the cross-device FL setting, where the parameter server does not perform user sub-sampling.

Lastly, we assume the standard, centralized, communication topology of FL as in real-world applications~\cite{gboard0, gboard1}, where each user is authenticated by a PKI and shares an encrypted channel with the server~$S$. The SA protocols in~\cite{bonawitz2017practical, bell2020secure} are designed for this communication topology.
%Moreover, note that a PKI is necessary in order to achieve security against a malicious server. 
Moreover, note that a PKI is a necessary assumption. Indeed, as described in~\cite[Section 6.2]{bonawitz2017practical} and~\cite[Section 4.2]{bell2020secure}, without a PKI, a server could break the privacy of users by simply launching a sybil attack.
Assuming the existence of a PKI is enough to rule out such trivial sybil attacks~\cite[Section 6.2]{bonawitz2017practical} and~\cite[Section 4.2]{bell2020secure}.

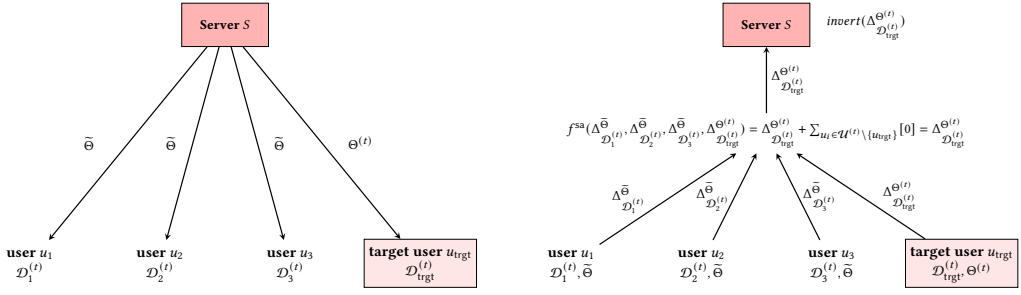
\begin{figure*}[th]
		\centering
		\begin{subfigure}{.4\textwidth}
			\centering
			\resizebox{.9\textwidth}{!}{
			\begin{tikzpicture}

				\tikzstyle{agg} = [rectangle, rounded corners, minimum width=1cm, minimum height=1cm,text right, draw=black, fill=white!30]
				\tikzstyle{server} = [rectangle, minimum width=2cm, minimum height=1cm,text centered, draw=black, fill=red!30]
				\tikzstyle{client} = [rectangle, minimum width=1cm, minimum height=1cm,text centered]
				\tikzstyle{target} = [client, fill=red!10, draw=black]
				\tikzstyle{arrow} = [thick,->,>=stealth]
				\tikzstyle{gradient}  = [line width=1.5mm, green!40, opacity=.8,]

				%\node (sa)[rectangle, rounded corners, minimum width=10cm, minimum height=4cm,text centered, draw=black, fill=white, opacity=.3, yshift=4.5cm, xshift=-.5cm, label=above: Server] {};

				\node (server) [server, yshift=0, xshift=-0cm] {\textbf{Server} $S$};

				\node (c0) [client,  below of=server, yshift=-4.5cm, xshift=-4.5cm] {\makecell[c]{\large{\textbf{user} $u_1$}\\$\dataset^{(t)}_1$}};
				\node (c1) [client,  right of=c0, xshift=2cm] {\makecell[c]{\large{\textbf{user} $u_2$}\\$\dataset^{(t)}_2$}};
				\node (c2) [client,  right of=c1, xshift=2cm] {\makecell[c]{\large{\textbf{user} $u_3$}\\$\dataset^{(t)}_3$}};
				\node (target) [target,  right of=c2, xshift=2cm] {\makecell[c]{\large{\textbf{target user} $u_\labeltarget$}\\$\dataset^{(t)}_\labeltarget$}};

				\draw [arrow] (server) -- (c0) node[midway, xshift=-.9cm] {$\widetilde{\Theta}$};
				\draw [arrow] (server) -- (c1) node[midway, xshift=-.5cm] {$\widetilde{\Theta}$};
				\draw [arrow] (server) -- (c2) node[midway, xshift=+.5cm] {$\widetilde{\Theta}$};
				\draw [arrow] (server) -- (target) node[midway, xshift=+.9cm] {$\Theta^{(t)}$};

		\end{tikzpicture}
	}

\end{subfigure}~\begin{subfigure}{.4\textwidth}
	\centering
	\resizebox{.9\textwidth}{!}{
	\begin{tikzpicture}

		\tikzstyle{agg} = [rectangle, rounded corners, minimum width=1cm, minimum height=1cm,text right, draw=black, fill=white!30]
		\tikzstyle{server} = [rectangle, minimum width=2cm, minimum height=1cm,text centered, draw=black, fill=red!30]
		\tikzstyle{client} = [rectangle, minimum width=1cm, minimum height=1cm,text centered]
		\tikzstyle{target} = [client, fill=red!10, draw=black]
		\tikzstyle{arrow} = [thick,<-,>=stealth]
		\tikzstyle{gradient}  = [line width=1.5mm, green!40, opacity=.8,]

		%\node (sa)[rectangle, rounded corners, minimum width=10cm, minimum height=4cm,text centered, draw=black, fill=white, opacity=.3, yshift=4.5cm, xshift=-.5cm, label=above: Server] {};

		\node (server) [server, yshift=0, xshift=-0cm] {\textbf{Server} $S$};

		\node (aggregator) [below of=server, yshift=-1.5cm, xshift=0] {$f^\labelSecAggr(\Delta^{\widetilde{\Theta}}_{\dataset^{(t)}_1},\Delta^{\widetilde{\Theta}}_{\dataset^{(t)}_2},\Delta^{\widetilde{\Theta}}_{\dataset^{(t)}_3},\Delta^{\Theta^{(t)}}_{\dataset^{(t)}_\labeltarget}) = \Delta^{\Theta^{(t)}}_{\dataset^{(t)}_\labeltarget} + \sum_{u_i \in \cUsers^{(t)} \setminus \{u_\labeltarget\}} [0] = \Delta^{\Theta^{(t)}}_{\dataset^{(t)}_\labeltarget}$};

		\node (c0) [client,  below of=aggregator, yshift=-2cm, xshift=-4.5cm] {\makecell[c]{\large{\textbf{user} $u_1$}\\$\dataset^{(t)}_1, \widetilde{\Theta}$}};
		\node (c1) [client,  right of=c0, xshift=2cm] {\makecell[c]{\large{\textbf{user} $u_2$}\\$\dataset^{(t)}_2, \widetilde{\Theta}$}};
		\node (c2) [client,  right of=c1, xshift=2cm] {\makecell[c]{\large{\textbf{user} $u_3$}\\$\dataset^{(t)}_3, \widetilde{\Theta}$}};
		\node (target) [target,  right of=c2, xshift=2cm] {\makecell[c]{\large{\textbf{target user} $u_\labeltarget$}\\$\dataset^{(t)}_\labeltarget, \Theta^{(t)}$}};

		\node (inversion) [client,  right of=server, xshift=1.3cm]{$invert(\Delta^{\Theta^{(t)}}_{\dataset^{(t)}_\labeltarget})$ };

		\draw [arrow] (server) -- (aggregator) node[midway, xshift=.5cm] { $\Delta^{\Theta^{(t)}}_{\dataset^{(t)}_\labeltarget}$};
		\draw [arrow] (aggregator) -- (c0) node[midway, xshift=-.9cm] {$\Delta^{\widetilde{\Theta}}_{\dataset^{(t)}_1}$};
		\draw [arrow] (aggregator) -- (c1) node[midway, xshift=-.5cm] {$\Delta^{\widetilde{\Theta}}_{\dataset^{(t)}_2}$};
		\draw [arrow] (aggregator) -- (c2) node[midway, xshift=+.5cm] {$\Delta^{\widetilde{\Theta}}_{\dataset^{(t)}_3}$};
		\draw [arrow] (aggregator) -- (target) node[midway, xshift=+.9cm] { $\Delta^{\Theta^{(t)}}_{\dataset^{(t)}_\labeltarget}$};

	\end{tikzpicture}
}

\end{subfigure}
\caption{Graphical representation of the gradient suppression attack (against FedSGD) with $\cUsers^{(t)}\myeq\{u_1, u_2, u_3, u_4 \myeq u_\labeltarget\}$.
The left figure depicts the malicious parameters distribution (model inconsistency).
The right figure depicts the secure aggregation of model updates, the collection of the target's model update, and the inversion.
The malicious parameters $\widetilde{\Theta}$ produce a tampered model update $\Delta^{\widetilde{\Theta}}_{\dataset^{(t)}_i}$ (i.e., gradient) equal to $[0]$ for each non-target user $u_i \in \cUsers^{(t)} \setminus \{u_\labeltarget\}$.
The function $invert(\cdot)$ denotes the technique used by the server $S$ (e.g., gradient inversion) to extract sensitive information of the original dataset $\dataset^{(t)}_\labeltarget$ from the target gradient $\Delta^{\Theta^{(t)}}_{\dataset^{(t)}_\labeltarget}$.}

% \caption{Graphical representation of the gradient suppression attack (against FedSGD) with $\cUsers^{(t)}\myeq\{u_1, u_2, u_3, u_4 \myeq u_\labeltarget\}$.
% 	The left figure depicts the malicious parameters distribution (model inconsistency).
% 	The right figure depicts the secure aggregation of users' model updates, the collection of the target's model update, and the inversion.
% 	The malicious parameters $\widetilde{\Theta}$ produce a tampered model update equal to $[0]$ for each non-target user $u_i \in \cUsers^{(t)} \setminus \{u_\labeltarget\}$.}
\label{fig:attack}
\end{figure*}

\section{Gradient suppression attack}
\label{sec:attack1}

In this section, we present our first attack, dubbed \textit{gradient suppression}, in which a malicious server exploits the model inconsistency attack vector to bypass SA and leak the model update of a chosen target user.

In a nutshell, the malicious server $S$ selects a \textbf{target user} $u_\labeltarget$ among the set of active users $\cUsers^{(t)}$ of the current round $t\in\NN$.
The aim of $S$ is then to preserve the target's model update during the SA process by tampering with the parameters for the other \textbf{non-target users}.
Here, $S$ creates a set of malicious parameters~$\widetilde{\Theta}$ that is sent to the non-target users $\cUsers^{(t)} \setminus \{u_\labeltarget\}$ whereas $u_\labeltarget$ receives the real parameters vector $\Theta^{(t)}$.
The malicious $\widetilde{\Theta}$ is crafted in such a way that the local application of gradient descent performed by a non-target users  produces a tampered model update $\Delta^{\widetilde{\Theta}}_{\dataset^{(t)}_i}$.
The tampered model updates, when aggregated through SA, have the property of preserving the target's model update $\Delta^{\Theta^{(t)}}_{\dataset^{(t)}_\labeltarget}$, allowing the server to recover it.
Once that the target's model update is on the server-side, $S$ can leak sensitive information about the batch $\dataset^{(t)}_\labeltarget$, used during the current round $t$ of FL by executing an arbitrary gradient inversion attack (see Section~\ref{sec:gradient-inversion-fl}) or related inference attacks.

Given the fact that the SA performs the sum among the users' model updates, the simplest way to achieve the isolation of the target's signal is to \textbf{force the tampered model updates to have a negligible magnitude, or, more strictly, to be zero everywhere}.
In this section, we first study the extreme case $\Delta^{\widetilde{\Theta}}_{\dataset^{(t)}_i}\myeq[0]$.
That is, the aggregation of the model updates (i.e., gradients) is equal to $u_\labeltarget$'s model update, i.e.,
\begin{align*}
	& f^\labelSecAggr(\Delta^{\widetilde{\Theta}}_{\dataset^{(t)}_1},\ldots, \Delta^{\widetilde{\Theta}}_{\dataset^{(t)}_{i-1}}, \Delta^{\Theta^{(t)}}_{\dataset^{(t)}_\labeltarget}, \Delta^{\widetilde{\Theta}}_{\dataset^{(t)}_{i+1}}, \ldots, \Delta^{\widetilde{\Theta}}_{\dataset^{(t)}_n}) \\
	&\quad = f^\labelSecAggr([0],\ldots, [0], \Delta^{\Theta^{(t)}}_{\dataset^{(t)}_\labeltarget}, [0], \ldots, [0]) = \Delta^{\Theta^{(t)}}_{\dataset^{(t)}_\labeltarget}
\end{align*}
allowing $S$ to exactly recover $\Delta^{\Theta^{(t)}}_{\dataset^{(t)}_\labeltarget}$.
\noindent
\Cref{fig:attack} depicts the gradient suppression attack against FedSGD. We stress that the attack applies to FedAVG as we will discuss in Section~\ref{sec:attack1-exec}.

Next, we show how to compute the malicious parameters $\widetilde{\Theta}$ required to perform the gradient suppression attack.
%\ifconference
We focus on the most widely adopted class of deep learning models---the one based on the \relu~activation function. \textbf{However, in Appendix~\ref{sec:gsea}, we show that our approach extends to arbitrarily composed architectures.}
%\fi

\subsection{Gradient suppression for \relu~layers: The dead-layer trick}\label{sec:dead-layer-trick}
The \textit{Rectified Linear Unit} (\relu) activation function:
\begin{equation}
	\relu(x) =
	\begin{cases}
		x & \text{if } x > 0\\
		0& \text{if } x \leq 0
	\end{cases}
	\label{eq:relu}
\end{equation}
is one of the core technical improvements that led to deep learning~\cite{relu1}.
Nowadays, this function is ubiquitous in computer vision architectures, representing the core building block of highly successful and standardized models such as ResNet~\cite{resnet}, DenseNet~\cite{densenet} and many others.
Outside the computer vision domain, the \relu~activation function is currently finding its place in Natural Language Processing (NLP) applications thanks to the success of transformer networks~\cite{attention, gpt2, gpt3}.

The \textit{dying-ReLU} problem~\cite{dyingrelu} is a phenomenon that naturally occurs during the training of deep neural networks that rely on the \relu~activation function.
When a layer $\ell$ \TT{dies}, it enters a state where it can only produce a constant output. More importantly, the dead layer $\ell$ cannot produce any gradient during the gradient descent iterations, i.e., the derivatives of its trainable parameters are zero regardless of the given input and loss function.

Despite the \textit{dying-\relu} phenomenon can naturally occur, we show that it can also be intentionally induced by a malicious server to prevent a network from producing a gradient for one or more sets of parameters.
Next, we describe how this can be achieved and exploited to perform our gradient suppression attack.

\subsubsection{Triggering the \textit{Dying-ReLU} phenomenon with malicious parameters}
The \textit{Dying-\relu} phenomenon is due to the piece-wise non-differentiability of the \relu~activation function.
Consider a neural layer $\ell$ with a $\relu$ activation function, i.e., $\ell(x) \myeq \relu(x \otimes \theta + b)$.
From~\Cref{eq:relu}, we can see that the \relu~function behaves as a constant function $\relu(x)=0$, whenever the input $x$ is equal or less than zero.
Since the derivative of a constant function is always $0$, we can easily conclude that, for any loss function $\loss$, then we have
$\frac{\partial\loss}{\partial \theta} = [0]$ and $\frac{\partial\loss}{\partial b} = [0]$ when $x \otimes \theta + b \leq 0$.
In other words, the layer $\ell$ receives zero gradient for its trainable parameters $\theta, b \in \Theta$ every time its pre-activation (\ie $x \otimes \theta + b$) is less or equal to zero.

A malicious server $S$ can exploit the above behavior of the~\relu\ function to kill a layer $\ell$ of a neural network~$f$, i.e., by forcing the pre-activation $x \otimes \theta + b$ of $\ell$ to be less or equal to zero.
This can be accomplished (without control over the input $x$ of an user) by computing some malicious trainable parameters $\widetilde{\theta},\widetilde{b} \in \widetilde{\Theta}$ of the layer $\ell$.

In more detail, the operator $\otimes$ (see Section~\ref{sec:nn}) is generally based on a multiplication-like operation between the input $x$ and the kernel $\theta$.
Therefore, we can easily force the pre-activation to be $[0]$ for any input~$x$ by just choosing $\widetilde{\theta}=[0]$ and $\widetilde{b}=[\mathbb{R}_{\leq0}]$.
Alternatively, having some knowledge on the input~$x$, we can rely on different setups for $\widetilde{\theta}$ and $\widetilde{b}$.
For instance, if $x$ is strictly positive (\eg because~$x$ is the output produced by a previous \relu-layer or because of the adopted input normalization process), it is enough to produce a malicious
$\widetilde{\theta}$ with negative numbers.
Instead, if a bound on $x$ is known (\eg $x\in[-1,1]$), we can just set the malicious bias vector $\widetilde{b}$ to a large enough negative number (e.g., fix $\widetilde{\theta}$ and set $\widetilde{b} = -max(\widetilde{\theta} \bigotimes x)$).

Now, an attacker can exploit  the \textit{dead-layer trick}  to force a \relu-based network to produce zero gradient for every layer.
In this direction, it is important to note that, for plain, \relu-based feedforward architectures, the server $S$ can just kill the kernels in the very first layer to suppress the gradient flow for the rest of the network.\footnote{However, if present, all the bias terms of the network should be set to values $\leq 0$.} Hence, killing all the kernels $\{\theta_i\}$ in the original parameters~$\Theta$ is very often not necessary.
Similarly, in modern architectures, neural layers tend to be arranged in the form:
	\[
	\text{neural layer} \rightarrow \text{normalization layer} \rightarrow \text{activation}.
	\]
	Therefore, to suppress the gradient for the network, it is enough to zero only the parameters of the normalization layers as these are often defined as $\gamma \hat{x}+\beta$, where $ \hat{x}$ is the normalized input.
	For instance, in the case of batch normalization, the attacker can kill the neural layer by setting the vectors ${\gamma}$ and ${\beta}$ of the batch normalization to $[0]$. Nevertheless, the strategy would work for every architectural configuration.
 
	Moreover, even if \relu~is the most common activation function in deep learning, a server $S$ can always maliciously choose a neural network architecture $f$ that presents a \relu~activation function in the ``right spots'' of the model without requiring unrealistic architecture modifications. The only gradient signal that cannot be recovered by the server using the dead-\relu~trick is the one of the bias term of the last layer 	(details are given in Appendix~\ref{sec:gsea}). This follows from the fact that the terminal layer of a network only rarely exhibits a \relu~activation function. A trivial solution for the malicious server is to avoid the bias term of the last layer when defining the architecture of the model. Alternatively, the malicious server can ignore the gradient of the bias term during the gradient inversion.
		Indeed, this represents only a tiny portion of the total number of trainable parameters of the network. For instance, in the case of a ResNet50 trained on ImageNet~\cite{resnet}, the bias vector in the final layer counts for only $4\cdot10^{-5}\%$ of the total number of parameters. In Section \ref{sec:att_s2}, we show that gradient inversion is unaffected from this missing gradient.
\begin{figure}[t]
	\centering
	\fboxsep=0.01mm%padding thickness
	\fboxrule=0mm%border thickness
	\begin{subfigure}{.8\linewidth}
		\centering
		\fcolorbox{black!50}{black!50}{\includegraphics[trim = 23mm 2mm 18mm 2mm, clip, width=.9\linewidth]{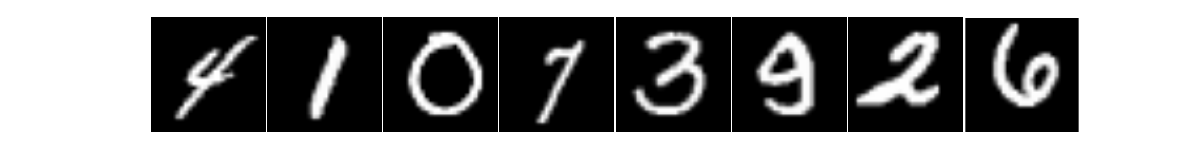}}\\
		\fcolorbox{red!50}{red!50}{\includegraphics[trim = 23mm 2mm 18mm 2mm,, clip, width=.9\linewidth]{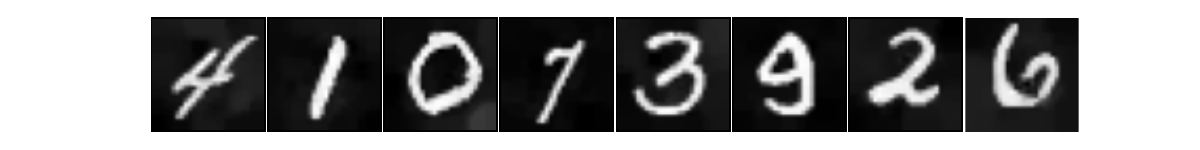}}
		\caption{MNIST}
	\end{subfigure}	\\

	\begin{subfigure}{.8\linewidth}
		\centering
		\fcolorbox{black!50}{black!50}{\includegraphics[trim = 23mm 2mm 18mm 2mm, clip, width=.9\linewidth]{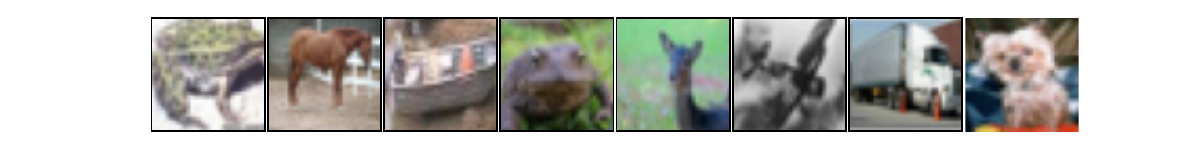}}\\
		\fcolorbox{red!50}{red!50}{\includegraphics[trim = 23mm 2mm 18mm 2mm, clip, width=.9\linewidth]{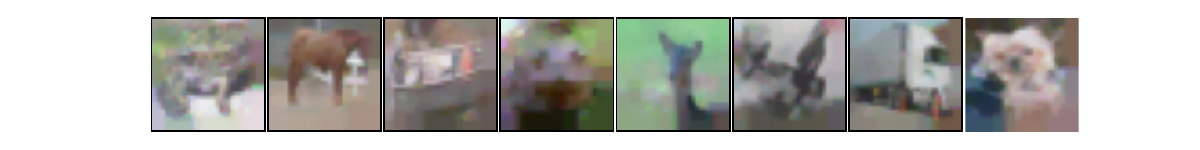}}
		\caption{CIFAR-10}
	\end{subfigure}	\\
	
	\caption{Examples of reconstruction (red panels) obtained via gradient inversion attacks using~\cite{invg2} for two datasets.}
	\label{fig:gi_example}
\end{figure}
\subsection{Attack execution}\label{sec:attack1-exec}
Turning back to the attack described at the beginning of this section, we have now an effective and efficient approach to isolate the model update $\Delta^{\Theta^{(t)}}_{\dataset^{(t)}_\labeltarget}$ of the target user~$u_\labeltarget$.
The malicious server $S$ can just exploit the techniques discussed in Section~\ref{sec:dead-layer-trick} and Appendix~\ref{sec:gsea} to generate the malicious parameters $\widetilde{\Theta}$ and suppress the model updates of non-target users, completely nullifying their contributions in the aggregated signal produced by SA.
As previously described (\Cref{fig:attack}), the attack is composed of two phases.

\subsubsection{Distribution of the (malicious) parameters}
In the first phase of the attack, $S$ creates the malicious parameters $\widetilde{\Theta}$ for the non-target users.
Right after the choice of $\widetilde{\Theta}$, $S$ must choose the target user $u_\labeltarget$ for the current round $t \in \NN$ of FL.
$S$ can either select the target at random (a trawling attack) from $\cUsers^{(t)}$ or target a specific (e.g., exploiting the IP address used to query the model by the user, if available).
Then, $S$ can enforce the model inconsistency by distributing the parameters (\Cref{fig:attack}).
In more detail, upon receiving a request for the parameters from a user $u_i$, the parameter server answers by sending $\Theta^{(t)}_i$ defined as follows:
\begin{equation}\label{eq:parameters-model-inconstency}
	\Theta^{(t)}_i =
	\begin{cases}
		\Theta^{(t)} & \text{if } i = \labeltarget\\
		\widetilde{\Theta} & \text{otherwise}
	\end{cases},
\end{equation}
where $\Theta^{(t)}$ are the honest parameters of the current round $t \in \NN$.
Optionally, $S$ can send a maliciously crafted model to the target user to increase the information recovered from the inversion attack~\cite{invg2, robbing}.
% Optionally, the malicious server can send an ill-conditioned model to the target user to artificially strengthen the gradient signal and help the inversion process. For instance, this can be achieved by performing a few clear iterations of distributed training in order to achieve a model with some utility and then applying a random perturbation on the rows of kernels of the last layer~\cite{invg2}.
\iffull
We stress that this procedure is agnostic to the used gradient inversion technique and that privacy leakage induced will increase with the improved gradient inversion techniques.
\fi

\subsubsection{Aggregation, collection, and inversion}
\label{sec:att_s2}
After the distribution of the parameters, the malicious server $S$ waits until it receives the output $v$ of $f^{\labelSecAggr}$, i.e., $f^\labelSecAggr(\Delta^{\widetilde{\Theta}}_{\dataset^{(t)}_1},\ldots,\Delta^{\Theta^{(t)}}_{\dataset^{(t)}_\labeltarget}, \ldots, \Delta^{\widetilde{\Theta}}_{\dataset^{(t)}_n}) = v$ where
\begin{align}\label{eq:secure-aggregation-attack}
	& v = \Delta^{\Theta^{(t)}}_{\dataset^{(t)}_\labeltarget} + \sum_{u_i \in \cUsers^{(t)} \setminus \{u_\labeltarget\}} \Delta^{\widetilde{\Theta}}_{\dataset^{(t)}_i}.
\end{align}
Then, it proceeds differently according to which algorithm (between FedSGD or FedAVG) is active.

In FedSGD, the output $v$ of $f^{\labelSecAggr}$ is the $u_\labeltarget$'s gradient, i.e., $v \myeq \Delta^{\Theta^{(t)}}_{\dataset^{(t)}_\labeltarget}\allowbreak  \myeq \nabla^{\Theta^{(t)}}_{\dataset^{(t)}_\labeltarget}$.
This is because, as discussed earlier, the malicious parameters $\widetilde{\Theta}$ produces $\Delta^{\widetilde{\Theta}}_{\dataset^{(t)}_i} \myeq [0]$ for each non-target user $u_i \in \cUsers^{(t)} \setminus \{u_\labeltarget\}$ (\Cref{sec:dead-layer-trick} and~\Cref{sec:gsea}).
After recovering the plaintext gradient, the server can reconstruct the target input by performing standard inversion attacks (Section~\ref{sec:gradient-inversion-fl}) as done in the protocol without SA. Figure~\ref{fig:gi_example} reports examples of gradient inversion on the federated system with gradient recovered via the gradient suppression attack. In the examples, we keep the bias term of the last layer in the architecture and ignore it during the optimization~\cite{invg2}. Note that the attack is independent of the number of users participating in the aggregation, and this can be arbitrarily large. Similarly, the server can perform previously proposed inference attacks~\cite{8835245, melis2019exploiting} on the individual user.

On the other hand, in FedAVG, a model update is composed of the parameters of the local model rather than a gradient (see~\Cref{sec:federate-learning}).
More formally,
\begin{align}\label{eq:update-fedSGD}
	\Delta^{\Theta^{(t)}_i}_{\dataset^{(t)}_i} &= \Theta^{(t,k)}_i & \text{ for } u_i \in \cUsers^{(t)},
\end{align}
where the local model $\Theta^{(t,k)}_i$ of $u_i$ is obtained by applying $k$ iterations of SGD using the local dataset $\dataset_i$ and $\Theta^{(t)}_i$ (as defined~\Cref{eq:parameters-model-inconstency}) sent by server (recall that $\Theta^{(t)}_i \myeq \widetilde{\Theta}$ for $\cUsers^{(t)} \setminus \{u_\labeltarget\}$).
Now, when a non-target user performs the local training procedure using the malicious parameters $\Theta^{(t)}_i \myeq \widetilde{\Theta}$, we have that
\begin{equation}\label{eq:stable-update}
	\Theta^{(t,j+1)}_i = \Theta^{(t,j)}_i - \eta \cdot \nabla^{\Theta^{(t,j)}}_{\dataset^{(t,j)}_i}
\end{equation}
where $\Theta^{(t,1)} \myeq \widetilde{\Theta}$, $\dataset^{(t,j)}_i \subseteq \dataset_i$, and $j \in \{1,\ldots,k\}$.
As in the case of FedSGD, we have that $\nabla^{\Theta^{(t,j)}}_{\dataset^{(t,j)}_i} \myeq [0]$ for every non-target user $u_i \in \cUsers^{(t)} \setminus \{u_\labeltarget\}$ and we conclude that $\Theta^{(t,j)}_i = \widetilde{\Theta}$.

By combining~\Cref{eq:secure-aggregation-attack,eq:update-fedSGD,eq:stable-update}, we obtain the equality
$v = (n-1)\cdot \widetilde{\Theta} + \Theta^{(t,k)}_\labeltarget$.
This equation can be solved with respect to the indeterminant $\Theta^{(t,k)}_\labeltarget$.
Once the malicious server $S$ recovered the updated local model $\Theta^{(t,k)}_\labeltarget$ of the target $u_\labeltarget$, it can determine the (pseudo) gradient signal by removing the honest parameters $\Theta^{(t)}$ from $\Theta^{(t,k)}_\labeltarget$ and proceed with the gradient inversion/inference attack.

\subsection{Impact}\label{sec:impact1}

Current FL+SA implementations do not prevent the gradient suppression attack, making users actively susceptible to this simple yet powerful exploit.
To a certain extent, this attack can be interpreted as an \textbf{invalid input validation vulnerability} present in the users' FL client software.
Here, the latter permits users to perform computation on ``semantically malformed inputs'' sent by a non-trusted party, i.e., the server. This allows the server to control SA's inputs of users and eventually affect the aggregation.
Furthermore, in contrast to most of the previous attacks introduced in FL, the disclosed vulnerability has the practical advantage \textbf{of being completely independent of the number of users participating in the current round.}
Therefore, this procedure scales to millions of active users, making it applicable to real-world scenarios such as cross-device FL, which is currently being deployed in-the-wild~\cite{gboard0,gboard1}. In the same direction, its effectiveness is independent of the size of the model or other nuisance factors such as the stillness of users' training datasets during the attack~\cite{gdis}. Additionally, unlike~\cite{robbing}, this attack neither hinges on auxiliary information on the users' private sets nor requires unrealistic modifications of the model architecture; indeed, it can be applied to arbitrary architectures and loss functions.
It is important to note that the gradient suppression attack can be iterated several times and arbitrarily alternated with honest training iterations. If the server wants to recover information on all the users, it has to iterate the attack several times by targeting a user at a time.
Assuming no dropouts among users participating at the FL protocol, recovering the gradient of all users requires $n$ iterations where $n$ is the number of active users.

We stress that the gradient suppression attack shows the incorrect application of SA in FL, yielding a ``false sense of security''.
%Indeed, the gradient attack does not assume any particular implementation of SA, and it works even if we replace the SA protocol with the ideal functionality $f^\labelSecAggr$.
As discussed in~\Cref{subsec:contribution}, the core motivation is that SA guarantees that nothing is leaked about the model updates of the users except what can be inferred from their aggregation.
\textbf{This claim assumes that the inputs (i.e., model updates) of SA are fixed and are not under the control of an adversary.
However, this does not hold in FL since, in this case, the value of inputs that needs to be aggregated depends on the parameters $\Theta$ distributed by the server.}
Hence, a malicious server that executes the gradient suppression attack can indirectly tamper with the SA's inputs to maximize the information leaked (e.g., leak the model update of a target user).

Although the gradient suppression attack is highly effective, it can be easily detected by non-target users (we delve into this topic in~\Cref{sec:def}). 
Nevertheless, in the next Section~\ref{sec:attack2} we introduce an extension of the gradient suppression attack that adds stealthiness and it is harder to detect. More generally, one can always trade effectiveness for stealthiness also in the gradient suppression framework. For instance, if recovering noisy model updates is acceptable, the server can send highly optimized models (\eg obtained after some rounds of honest execution) to non-targets and an unoptimized model to the target. Intuitively, the gradient from the unoptimized model should dominate the aggregation given the low magnitude and sparsity of the one produced by the optimized models (see Figure~\ref{fig:sparsity}).% Similarly, in case of a \relu-based architecture, the server can manipulate both target and non-target's parameters by suppressing a disjointed set of kernels. That is, if $m$ is a binary mask defining which kernels of the target's model to kill, we suppresses the $\not m$ kernels in the non targets. In this case, the server is able to fully recover the target's gradient, but without forcing non-targets to produce zero-gradient everywhere (again, gradient sparsity is a natural phenomenon).}

\section{Canary-gradient attack for property inference}
\label{sec:attack2}
\Cref{sec:attack1} demonstrates that a malicious server $S$ can force non-target users to produce a zero gradient during a round of FL.
This allows $S$ to bypass SA and, at the same time, maximize the leakage regarding the dataset of a target user.
While the gradient suppression attack can be seen as the most extreme exploitation of the model inconsistency attack vector, more stealthy attacks can be created harnessing the same underlying intuition.

This section shows a general procedure that allows a malicious server to perform highly accurate property inference attacks on individual users, even if SA is enabled. The idea behind this approach is that the server can maliciously modify the parameters of the model in order to inject specific detectors in one or more subsets of the network. These detectors are specifically crafted to react to attacker-chosen trigger conditions that can be present in the users' training instances. Whether the detector is triggered during the local training procedure, the network produces a clear footprint in the model update. Then, upon receiving the latter from a user, the server can determine if the trigger condition has been met by looking for the footprint in the model update. This allows the server to infer information on the content of the user's training set; that is, the presence or absence of data with the specific property. For instance, using this approach, the server can perform an extremely accurate membership inference on a chosen target user. Hereafter, we refer to this general procedure as the \textbf{canary-gradient~attack}.

As for the gradient suppression, the canary-gradient attack does not target any specific SA protocol but instead leverages a vulnerability of FL caused by its incorrect usage of SA. For this reason, we abstract the SA protocol with its ideal functionality $f^\labelSecAggr$.
% Indeed, it succeeds even if FL uses a perfectly secure SA protocol, defined by the ideal functionality $f^\labelSecAggr$.
%We stress that, in this case, we only focus on FedSGD, i.e., the model update of a user is a gradient $\Delta^{\Theta^{(t)}}_{\dataset^{(t)}_i} = \nabla^{\Theta^{(t)}}_{\dataset^{(t)}_i}$.

\subsection{The Conditional Dead-Layer trick}\label{sec:conditioned-dead-layer-trick}
The main building block to construct the attack is a conditioned version of the dead-layer trick of~\Cref{sec:dead-layer-trick}.
Informally, we want to ``kill'' a layer only if the instance $x \in \dataset^{(t)}_\labeltarget$ of the target $u_\labeltarget$ satisfies a particular condition.
In other words, we would like a programmed death through the backdooring of the layer.
For the sake of presentation, we introduce the conditional dead-layer trick assuming that SA is disabled.
Then, in~\Cref{sec:target-property-inference} we extend the discussion to the case of SA enabled.

Formally, given a layer $\ell$, we want to find some malicious parameters $\widetilde{\Theta}$ to enforce the following behavior
\begin{equation}
	\label{eq:canary0}
	\frac{\partial \loss(\dataset^{(t)}_{\labeltarget}, \widetilde{\Theta})}{\partial \xi} \neq 0 \Longleftrightarrow \exists x \in \dataset^{(t)}_\labeltarget: P(x) \myeq \mathsf{True},
\end{equation}
where $\dataset^{(t)}_\labeltarget$ is the batch (of the current round $t$) used by the target user $u_\labeltarget$, $\xi \in \widetilde{\Theta}$ is a  subset parameters of  the network, and $P$ is a predicate that defines the property the malicious server $S$ wants to detect in the batch $\dataset^{(t)}_\labeltarget$ of $u_\labeltarget$.
In particular, $\xi$ can be composed of the parameters of any logic partition in the neural network, such as a specific filter in a convolution layer or an element in the scale and shift vectors in a normalization layer.

% In the setting of FL, in a given round $t$ of FL, the malicious parameters $\widetilde{\Theta}$ (that include $\xi$) will locally force the network $f_{\widetilde{\Theta}}$ to produce a gradient $\nabla^{\widetilde{\Theta}}_{\dataset^{(t)}_\labeltarget}$ for the user $u_\labeltarget$.
% In turn, $\nabla^{\widetilde{\Theta}}_{\dataset^{(t)}_\labeltarget}$ will include a nonzero gradient $\nabla^{\xi}_{\dataset^{(t)}_\labeltarget}$ for a subset of its parameters $\xi$ if and only if $u_\labeltarget$ has used a batch $\dataset^{(t)}_\labeltarget$ containing at least one instance $x\in\dataset^{(t)}_\labeltarget$ such that $P(x) = \mathsf{True}$ (\ie the backdoor trigger).
% This $\nabla^{\xi}_{\dataset^{(t)}_\labeltarget}$ will compose the footprint identifiable by $S$.

As discussed in Section~\ref{sec:attack1}, suppressing the gradient for a set of parameters in a \relu-based layer is about controlling the value of its pre-activation.
Therefore, given a neural layer $\ell$ with \relu~activation, we can substitute~\Cref{eq:canary0} with:
\begin{equation}
	\label{eq:canary1}
	\ell_{\xi}\myeq (x_{\xi} \otimes \theta_{\xi} + b_{\xi}) > 0 \Longleftrightarrow \exists x \in \dataset^{(t)}_\labeltarget: P(x) \myeq \mathsf{True},
\end{equation}
where  $\xi = \{\theta_{\xi},b_{\xi}\}$, and $x_{\xi}$ is the subset of the input of~$\ell$ that interacts with the parameters $\xi$ and $\ell_{\xi}$ refers to the subset of the output of the layer~$\ell$ computed using the parameters $\xi$.

The simplest and most natural way to find $\xi$ that correctly induce~\Cref{eq:canary1}
is to explicitly train the layers preceding $\ell$ and the parameters $\xi$ to force $\ell_{\xi}$ to produce a positive value only when the input of the network satisfies $P$.
In other words, we train part of the network in a classification task, using the output $\ell_{\xi}$ such as the output layer, where the classification threshold is centered in zero.
Observe that, if the behavior of~\Cref{eq:canary1} is correctly embedded in the network~$f_{\widetilde{\Theta}}$, a malicious server $S$ will able to determine the event $\exists x \in \dataset^{(t)}_\labeltarget: P(x) \myeq \mathsf{true}$ by only collecting gradient $\nabla^{\xi}_{\dataset^{(t)}_\labeltarget}$ of $u_\labeltarget$ and check that the derivatives of $\xi$ are different from zero.
In Section~\ref{sec:injection} we show how this can be done in practice.

\begin{figure}
	\begin{centering}
		\includegraphics[trim = 0mm 0mm 0mm 0mm, clip, width=.7\linewidth]{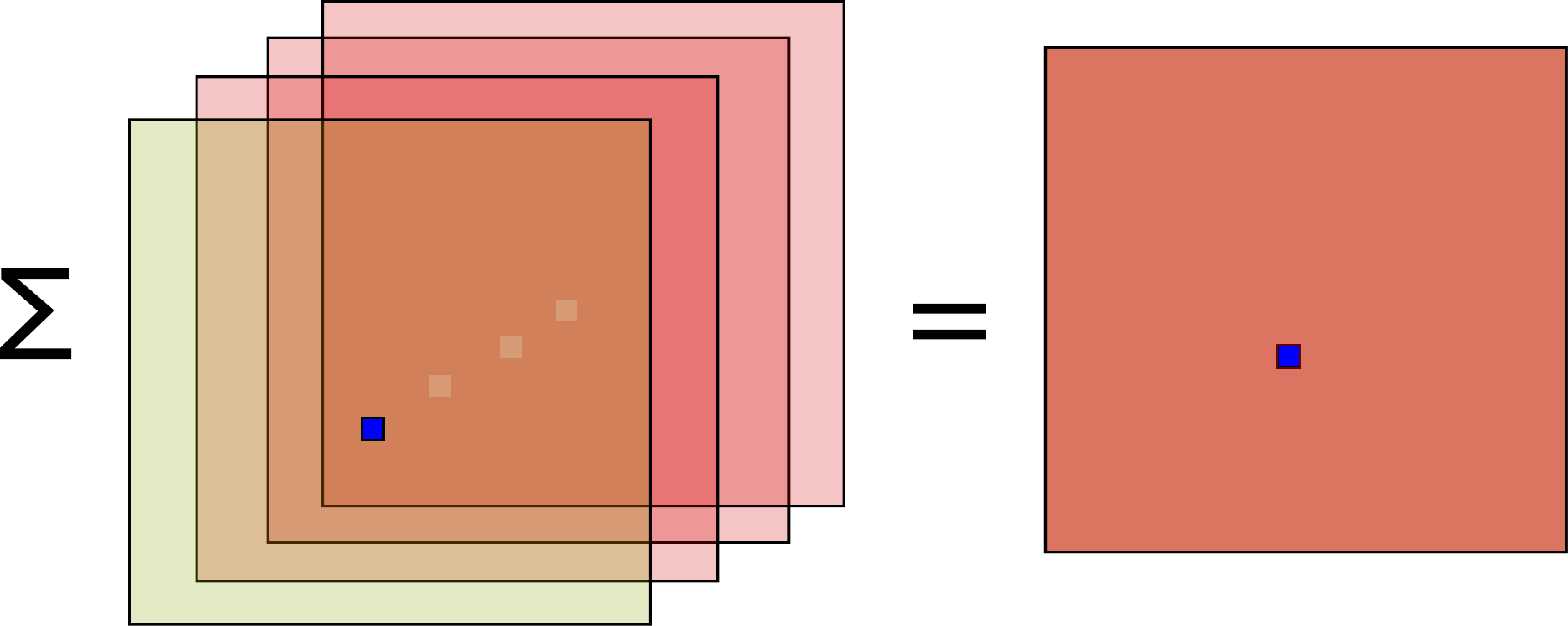}\\
		\caption{Graphical representation of the SA exeuction when the canary-gradient is applied.
	On the left, each square represents a gradient update produced by a different user.
	The green square represents the target's gradient and the inner small blue square represents the gradient for $\xi$.
	On the other hand, each red square represents the gradient produced by non-target users, with zero gradient for $\xi$.
	The square on the right represents the aggregation where the target's gradient $\nabla^{\xi}_{\dataset^{(t)}_
		\labeltarget}$ for $\xi$ is preserved.}
		\label{fig:canary}
	\end{centering}
\end{figure}

\begin{figure}
	\begin{centering}
		\includegraphics[trim = 0mm 0mm 0mm 0mm, clip, width=.7\linewidth]{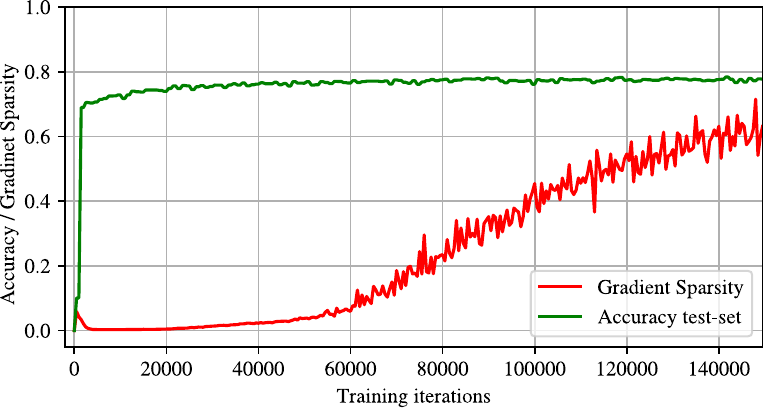}\\
		\caption{Comparison between the test-set accuracy of a ResNet model trained on CIFAR10 and the sparsity of its gradient (\ie percentage of parameters that receive zero gradient) during the training. %Here, the gradient becomes sparser and sparse as the training progresses.
		}
		\label{fig:sparsity}
	\end{centering}
\end{figure}

\subsection{Targeted property inference attacks via model inconsistency}\label{sec:target-property-inference}
To perform the membership inference attack discussed in the previous section, the malicious server $S$ needs to have access to the gradient $\nabla^{\xi}_{\dataset^{(t)}_\labeltarget}$ of $u_\labeltarget$.
This is possible when SA is disabled.
On the other hand, when SA is enabled, the $S$ can inject the canary-gradient functionality in the network and then perform the inference attack on the whole pool of active users.
% Then, $S$ only sees the following final aggregation
% \begin{align}\label{eq:sa-xi}
% 	&f^{\labelSecAggr}_{\xi}(\nabla^{\xi}_{\dataset^{(t)}_1},\ldots,\nabla^{\xi}_{\dataset^{(t)}_\labeltarget},\ldots, \nabla^{\xi}_{\dataset^{(t)}_n}) = \nonumber \\
% 	& \quad = \nabla^{\xi}_{\dataset^{(t)}_\labeltarget} + \sum_{u_i \in \cUsers^{(t)}\setminus\{u_\labeltarget\}} \nabla^{\xi}_{\dataset^{(t)}_i} = v,
% \end{align}
% where $f^{\labelSecAggr}_{\xi}$ denotes the inner idealized functionality of $f^\labelSecAggr$ that performs the aggregation of the gradients of the parameter $\xi$.
In this scenario, $S$ would be able to infer that one of the users triggered the canary-gradient by observing that $f^{\labelSecAggr}_{\xi}(\nabla^{\xi}_{\dataset^{(t)}_1},\ldots, \nabla^{\xi}_{\dataset^{(t)}_n}) = \sum_{u_i \in \cUsers^{(t)}} \nabla^{\xi}_{\dataset^{(t)}_i} \neq  [0]$ where $f^{\labelSecAggr}_{\xi}$ denotes the inner idealized functionality of $f^\labelSecAggr$ that performs the aggregation of the gradients of $\xi$.
However, in this case, the privacy of users would be partially preserved as $S$ would not be able to attribute the result of the inference attack to a specific user (\TT{privacy by shuffling}).

Still, we show that model inconsistency can be exploited even in this case, allowing the malicious server $S$ to bypass SA and target the specific target $u_\labeltarget$.
Analogously to the gradient suppression attack (\Cref{sec:attack1}), $S$ needs to tamper with the honest parameters $\Theta^{(t)}$ in order to produce two different malicious $\widetilde{\Theta}_1$ and $\widetilde{\Theta}_2$.
The target user $u_\labeltarget$ will receive $\widetilde{\Theta}_1$ that is the original model $\Theta^{(t)}$ injected with a canary-gradient for the parameters $\xi$ as discussed in~\Cref{sec:conditioned-dead-layer-trick}.
On the other hand, the non-target users $\cUsers^{(t)} \setminus\{u_\labeltarget\}$ will receive $\widetilde{\Theta}_2$ that is a slight perturbation of the original model $\Theta^{(t)}$ that has the additional property of unconditionally produce zero-gradient only for the parameters $\xi$, i.e., $\nabla^{\xi}_{\dataset^{(t)}_i} \myeq [0]$.
This can be achieved by exploiting the dead-layer trick in a localized way.
Instead of killing the gradient for the whole layer, it intentionally inhibits only the gradient produced by the parameters $\xi$; for instance, for just one filter in a convolution layer.

% As a consequence, we conclude that
% \[
% v = \nabla^{\xi}_{\dataset^{(t)}_\labeltarget},
% \]
% where $v$ is the output of $f^\labelSecAggr_\xi$ defined in~\Cref{eq:sa-xi}
Now, when the target and non-target gradients are aggregated, the target's gradient $\nabla^{\xi}_{\dataset^{(t)}_\labeltarget}$ for $\xi$ will be preserved, allowing the server to state the activation or non-activation of the canary-gradient, and so, with high probability, the presence of the property $P$ in the batch $\dataset^{(t)}_\labeltarget$ of $u_\labeltarget$.
This intuition is captured by~\Cref{fig:canary}.

Finally, given that the number of parameters in $\xi$ can be arbitrarily small, the attack will leave only a minimal footprint, making the detection non-trivial (in our experiments, we show that two parameters are enough).
This is also supported by the fact that the gradient becomes more sparse as the training proceeds (see Figure~\ref{fig:sparsity}), making it difficult to distinguish between natural \TT{holes} and artificial ones in the gradient vector.

While we gave an abstract view on the attack strategy, next, we show how the attack can be carried out on a realistic architecture such as ResNet in a membership inference attack scenario.

\subsection{Injecting canary-gradient for membership inference}\label{sec:injection}
We show a practical example of how to model a membership attack on the training dataset of a specific user.
Specifically, we target training instance $x_t$, and we want to infer if $x_t$ is contained in the batch $\dataset^{(t)}_{\labeltarget}$ used by the target user $u_\labeltarget$ to compute the gradient update in the current round $t \in \NN$ of FL.
Following previous notation, we want to infer the following property:
\begin{equation*}
	P_{x_t}(x) = \mathsf{True} \Longleftrightarrow x = x_t.
\end{equation*}
We start by considering the case of FedSGD and carry out the attack on a ResNet20 network.
However, for this network, we do not consider the batch normalization layers as those would make the attack trivial.
Indeed, if batch normalization is used, we could detect the activation of $\ell_\xi$ by checking the average computed and sent to the server to update the running mean.
For this reason, we keep the attack general by substituting every batch normalization with layer normalization~\cite{layernorm} that does not present this issue and has an overlapping role. We then extend the attack to FedAVG in Appendix~\ref{app:canary_avg}.

In our experiment, we inject the canary gradient in the last residual block of the network.
This is because the terminal layers are usually the ones that receive the sparsest gradient during the training.
Since the normalization layer precedes the \relu~activation function, we chose a subset of the parameters of the latter as our $\xi$.
In particular, we can pick any pair $(\gamma_i, \beta_i)$ in the scale and shift vectors $\gamma$ and $\beta$. Thus, in this case, $\xi$ is composed of only two parameters, that is about {$7 \cdot 10^{-4}\%$} of the total number of parameters in the network.
Hereafter, we always choose $i=0$; however, choosing a different channel would not affect the attack.

To inject the canary gradient, we use a learning-based approach.
In this direction, we assume that the adversary (i.e., malicious server) knows a shadow dataset $\dataset_s$ defined in the same domain of the target point $x_t$.
For instance, if $x_t$ is a face image, $\dataset_s$ contains face images as well.
We stress that, as we will show later, the distribution of $\dataset_s$ and one of the users' datasets can be different.
Intuitively, the role of $\dataset_s$ is providing negative samples while training $\ell_{\xi}$ to fire on $x_t$.
\textbf{It is important to note that the canary-gradient attack is agnostic, and the simple training procedure discussed can be substituted with other techniques.} For instance, approaches such as~\cite{robbing} and learning-based approaches that do not require negative samples (\ie one-class classification) may be used to reach the same result. In this direction, our main contribution is introducing the idea that the server can manipulate the derivative of the current model to encode arbitrary messages (and that these messages can be retrieved under SA). %We call this general class of attacks \TT{First order inference attacks}, as those are based on the derivative of the neural network rather than directly on the function $f_\Theta$ itself.}

\begin{figure}
	\begin{centering}
		\includegraphics[trim = 0mm 30mm 0mm 35mm, clip, width=.6\linewidth]{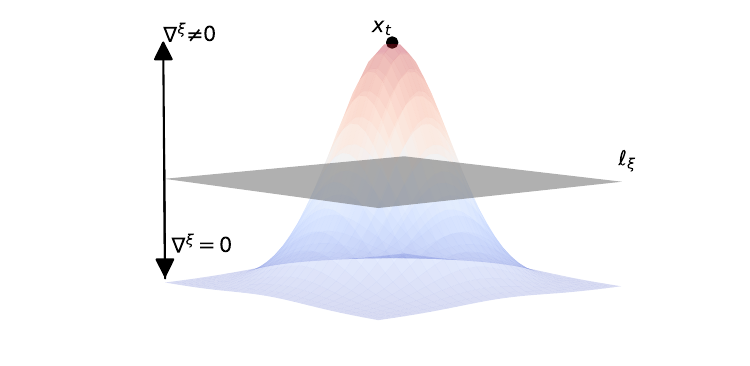}\\
		\caption{Graphical representation of the feature-space ($x$-axis and $y$-axis) for the pre-activation $\ell_\epsilon$ ($z$-axis). The gray plane represents $z=0$, i.e., the threshold for the activation of the \relu~function.}
		\label{fig:canary3d}
	\end{centering}
\end{figure}

We can now inject the canary gradient by reducing the malicious injection to a classification problem.
We train the split of the network up to $\xi$ in producing positive $\ell_\xi$ when the network's input contains $x_t$.
In our case, with $\xi=\{\gamma_i,\beta_i\}$, we have $\ell_\xi = \gamma_i\bar{x_i}+\beta_i$, where $\bar{x}$ is the normalized input of the layer.
Our loss function $\loss$ for a batch of size $n$ is simply defined as the binary cross entropy:
\begin{equation}
	-\frac{1}{n}\sum_{i=1}^n \begin{cases} \alpha_1 \cdot \log(\texttt{sigmoid}(\ell_\xi)) & \text{if } x = x_t \\
		\alpha_{0} \cdot \log(1-\texttt{sigmoid}(\ell_\xi)) & \text{otherwise}
	\end{cases},
\end{equation}
\iffalse
Our loss function $\loss$ for a batch of size $n$ is simply defined as:

\begin{equation}
	\frac{1}{n}\sum_{i=1}^n \begin{cases} \alpha_1 \cdot MSE(\ell_\xi, [+1]) & \text{if } x = x_t \\
		\alpha_{-1} \cdot MSE(\ell_\xi, [-1]) & \text{otherwise}
	\end{cases},
\end{equation}
\fi
where instances different from $x_t$ are sampled from $D_s$, and $\alpha_1$ and $\alpha_0$ weight the loss for the two events. % \new{Potentially, one can also set the bias term manually in the interval $[-1,0]$ to express a suitable threshold for the trigger activation as it would be done in any normal binary classification. In our approach}
In other words, we want to ``overfit'' the network to produce positive~$\ell_{\xi}$ only for the point~$x_t$, while squashing under~$0$ the feature-space around it.
This intuition is depicted in Figure~\ref{fig:canary3d}. We stop the training when we reach a training loss lower than a given threshold.% \new{(this is $0.0001$ for all the tested configurations)}.
\par

Finally, the gradient for $\xi$ in the non-target users is unconditionally suppressed by just setting both $\gamma_i = \beta_i = 0$, but other configurations are possible.

\begin{figure}[t]
	\begin{centering}
		\includegraphics[trim = 0mm 0mm 0mm 0mm, clip, width=1\linewidth]{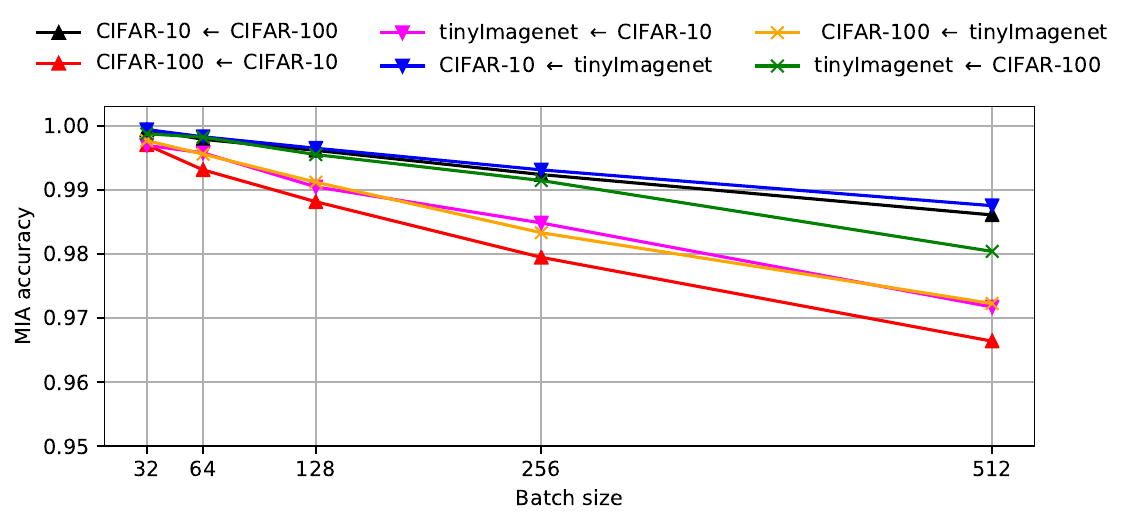}
		\caption{Average accuracy of the canary-gradient attacks for six different setups with increasing batch size for a ResNet20.}
		\label{fig:miaresult}
	\end{centering}
\end{figure}

\begin{figure}[t]
	\begin{centering}
		\includegraphics[trim = 10mm 0mm 9mm 0mm, clip, width=.5\linewidth]{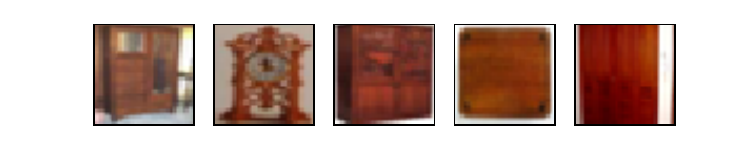}~\includegraphics[trim = 9mm 0mm 10mm 0mm, clip, width=.5\linewidth]{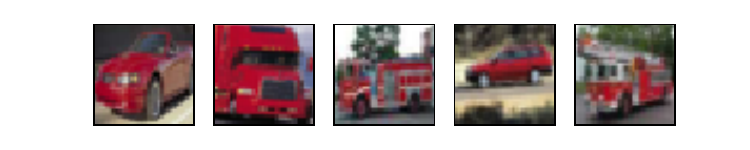}
		\\
		\includegraphics[trim = 10mm 3mm 9mm 0mm, clip, width=.5\linewidth]{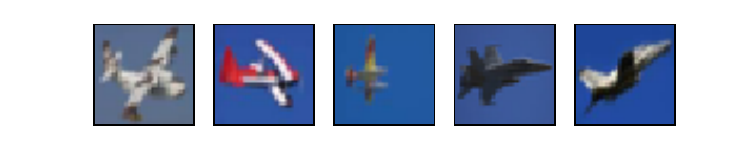}~\includegraphics[trim = 9mm 3mm 10mm 0mm, clip, width=.5\linewidth]{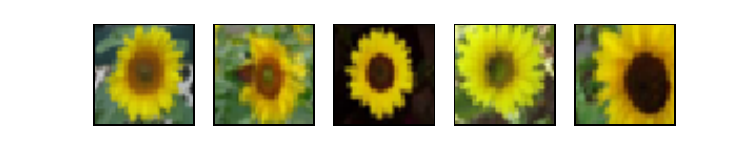}
		\caption{Four different examples of false positive for the canary-gradient-based MIA. The first element in each sequence is the target, whereas the four following images are false triggers.}
		\label{fig:false_positive}
	\end{centering}
\end{figure}
\subsubsection{Results}\label{sec:results}
We evaluated the effectiveness of our attack.
We test three different image datasets, namely, CIFAR10, CIFAR100~\cite{cifar}, and TinyImagenet~\cite{tiny}. We use all the possible permutations of those datasets to represent the private and shadow distribution for the canary-gradient attack, obtaining $6$ different configurations.

%For each dataset, we selected a different shadow dataset $\dataset_s$ defined in the same domain but with different distributions.
%To this end, we chose CIFAR10~\cite{cifar}, CelebA~\cite{celeb} and STL10~\cite{stl}.
%We selected the above shadow datasets since their distributions present lower entropy compared to the respective private ones (\eg CIFAR10 v.s. CIFAR100).
%This highlights the robustness of the attack against discrepancy in the private and shadow distributions.

To run the experiments, we pick $x_t$ (\ie the target of the membership inference) at random from the validation set of the private dataset.
Then, we trained the model by injecting the canary-gradient in the channel $0$ of the last normalization layer in the last residual block.
After the injection, we evaluated the canary's effectiveness by testing that the canary-gradient is non-zero when $x_t$ is in the training batch and zero otherwise.
To this end, we iterate over all the training data of the private dataset by selecting a batch $\cX$ of size $n$ at a time.
Given $\cX$, we compute the gradient according to the original loss function and test that $\xi$ has gradient zero (precision).
Then, we insert $x_t$ in $\cX$, and we test if the gradient of $\xi$ is different from zero (recall).
We perform the test on the three private datasets with different batch sizes and repeat the test for $50$ different $x_t$ for every case.We report our results in~\Cref{fig:miaresult}, where we use the notation \TT{\textit{private-dataset}$\leftarrow$\textit{shadow-dataset}}.

\textbf{The canary gradient has perfect recall} (\ie if $x_t$ is in the batch, the canary is always triggered), but it can be subject to false-positive errors with low probability.
The global accuracy of the method is about $99\%$ for a batch size up to $128$. The precision of the attacks slowly decreases when the batch size becomes larger.
This follows from the fact that larger batches have more probability of including at least one ``false trigger example'' that causes a false positive. Intuitively, false positives are instances that induce a feature representation that is similar to the target one; a few examples are given in Figure~\ref{fig:false_positive}. Therefore, while false triggers reduce the accuracy of the attack, these can still inform the attacker that instances \textit{similar} to $x_t$ are present in the victim's local training set. The attacker can intentionally tune the definition of similarity by introducing the desired inductive bias in the canary injection process.\footnote{For instance, the server can train the canary to intentionally react only to \TT{red cars}, by training it accordingly.}
Regardless of the presence of false triggers, the attack is appreciably precise; the accuracy remains higher than $96\%$ even in the worst configuration.
\textbf{The canary-gradient attack also applies to FedAVG} by just requiring minimal effort to the malicious parameter server. In Appendix~\ref{app:canary_avg}, we empirically demonstrate the effectiveness of this approach.

\subsection{Impact}
Although we presented the canary-gradient attack to execute a membership inference, the same approach can be applied to any type of property inference.
Ideally, it is sufficient to define a different trigger condition and train the network accordingly.
Moreover, the server can inject multiple canary-gradient with different triggers in the same network and infer non-binary properties.
In the same direction, the server can simultaneously perform inference on multiple users without losing accuracy by carefully managing the allocation of conditional and unconditional dead-layers.

More importantly, the canary-gradient attack maintains the same practical properties as the gradient-suppression attack; mainly, it requires only a training round to perform, its effectiveness is independent of the number of users participating in the round, and it is loss agnostic (\ie it works for any learning task).
However, unlike the gradient suppression approach, this leaves only a minimal footprint in the model updates. %, making the prediction non-trivial even for aware users.
Note that the canary gradient can be injected while training the model on the original task, and so minimizing the utility loss of the target model. On the other hand, suppressing a limited number of parameters in the non-target models (\eg two in our example) has only a negligible impact on the utility.\footnote{This is equivalent to perform dropout~\cite{dropout} on a single channel on a single layer in the network.}

This attack demonstrates that a malicious server can perform highly accurate property inference attacks on individual users even if SA is adopted from the latter. Again, the introduced training-based approach is just an example, and more sophisticated techniques may be devised to reach the same (or better) results by relying on the same general construction.
%However, while the attack is effective against FedSGD, this may have limited effectiveness on FedAVG.
%Indeed, in this case, the user changes the model parameters during the local training procedure, potentially overriding the gradient-canary functionality.
%In this direction, the attacker should devise robust canary-gradients (\eg similarly to robust backdoors~\cite{rbds}) that preserve functionality when altered via training.

For completeness, we emphasize that, even in the case of canary-gradient attacks, the discussion about the insecurity of the combination of SA and FL applies (discussed in~\Cref{subsec:contribution,sec:impact1}). As for the case of gradient suppression, the canary-gradient works if the SA protocol is perfectly secure (i.e., it behaves as the ideal functionality $f^\labelSecAggr$).

\section{Mitigations}\label{sec:def}
Next, we will discuss and introduce some mitigation approaches.

The first part focuses on heuristic approaches that strive to prevent model inconsistency and attack vectors similar to those seen in this work.
While they do not stop all possible attacks, their practical relevance is still significant.

Then, we analyze how combining DP with SA can lead to a more general mitigation strategy.

\subsection{Heuristic mitigations against model inconsistency}\label{sec:heuristic-mitigation}

\paragraph*{SA dropout}
Although the gradient suppression attack (Section~\ref{sec:attack1}) is effective, it can be detected easily by non-target users.
Indeed, non-target users can quickly discover an ongoing attack as their models would have zero gradients throughout.\footnote{A phenomenon that is very unlikely to be observed in reality, excluding numerical errors or pathologic overfitting.} However, the target (who is the victim of the attack) cannot detect or prevent the attack without communicating with non-target users.\footnote{Note that the standard implementation of FL does not allow users to communicate.}
Therefore, a solution could be instructing users not to execute the SA protocol when they detect a null gradient after a local training iteration.
In this case, only the attack victims would participate in the SA protocol, which is enough to prevent the server from recovering the target's model update.
Indeed, the server would not receive enough information to unmask the target model's update and complete the execution of SA.
This is because the dropout threshold $\delta$ of the underlying SA protocol (i.e., the number of user dropout that the SA protocol can handle) must be set to $\delta < n - 1$ where $n$ is the number of users participating in the aggregation (otherwise, if $\delta = n-1$, a malicious user can unmask the gradients of other users). See~\cite{bonawitz2017practical} for more details.
Nevertheless, this detection strategy does not work in the case of the canary-gradient attack (Section~\ref{sec:attack2}).
The ``negligible'' footprint of the model update does not allow for a reliable detection since sparse model updates are common in an honest execution of FL (see~\Cref{fig:sparsity}).

\paragraph*{Parameter validation using signatures}
Another approach consists of checking that all users have received the same parameters $\Theta^{(t)}$ in the current round $t \in \NN$ of FL.
For example, each $u_i \in \cUsers^{(t)}$ sends the parameters $\Theta^{(t)}_i$ (or its hash) sent by $S$ to all other users.
Then, each $u_i \in \cUsers^{(t)}$ checks that
\begin{equation}\label{eq:check}
	\Theta^{(t)}_i = \Theta^{(t)}_j \text{ for every } u_i,u_j \in \cUsers^{(t)}.
\end{equation}
If the check fails, the user aborts.
It is easy to see that this strategy does not allow $S$ to execute model inconsistency attacks  (note that it can still send a malicious parameters $\widetilde{\Theta}$ to all).
Unfortunately, in the standard communication topology of FL (see~\Cref{sec:threat-model}), users do not have a direct communication channel.
Each message needs to go through the server that, in turn, will forward the message to the intended receiver. Hence, a malicious server can perform a man-in-the-middle attack and substitute each $\Theta^{(t)}_i$ sent by $u_i$ with an arbitrary honest-looking parameters $\Theta$.
If we want to preserve the communication topology of FL, digital signatures must be involved.
For example, assuming that each $u_i$ holds a key pair $(sk_i,pk_i)$ then $u_i$ needs to $(i)$ sign its message before sending it to the server $S$ and, $(ii)$ verify the signature (using the public key $pk_j$ of the signee) of each message received.
If the security of the signature holds, then the server can not change the messages of users. As a consequence, users are guaranteed that the message is honest, and they can perform the check defined in~\Cref{eq:check}.
We stress that this approach works under the assumption that users $\cUsers$ can trust and know (in advance) the public keys of all other users (e.g., there is a trusted certification authority).
Note that this assumption is at the root of the SA protocols of Bonawitz et al.~\cite{bonawitz2017practical} and Bell et al.~\cite{bell2020secure}.

% Lastly, we stress that this approach adds one round of communication. This is particularly prohibitive in FL since users are volatile, and we would like to let them go offline as soon as possible.
% In some cases, this round can be amortized.
% For example, if the SA protocols of~\cite{bonawitz2017practical,bell2020secure} are used,
% we can merge this additional round of communication with ones of the setup of SA, i.e., the rounds required to compute the pairwise PRG's seeds (fundamental to execute the SA protocol).
% However, in several iterations of FL, the setup of SA is not executed (for example, when any pair of active users have never dropped out and already have a pairwise PRG’s seed. Hence, they can execute the SA protocol without re-running the setup).
% In these iterations, an additional round of communication is still required, affecting FL's efficiency.

This approach would add one round of communication without modifying the implementation of the underlying SA protocol. However, this round could be merged with the ones of SA, e.g., the third round of the SA protocol of Bonawitz et al.~\cite[Figure 2]{bonawitz2017practical}.
Regarding the communication complexity, exchanging the signatures of the received parameters increases communication by $n\cdot \ell$ where $n$ is the number of active users and $\ell$ is the size of one signature (e.g., $256$ bits).

\paragraph*{Conditional secure aggregation}
Another mitigation consists of building a modified SA version for FL that performs the aggregation only if a particular condition $C$ is satisfied.
Otherwise, it outputs a random value (or fixed value $\bot$ denoting that the aggregation did not occur).
Intuitively, by setting the condition $C$ to~\Cref{eq:check}, the FL protocol executes the aggregation (and continue its execution) only if all the users have received the same parameters $\Theta^{(t)}$ in the current round $t \in \NN$.
This would hinder a malicious server $S$ from exploiting the model inconsistency attack vector.
Naturally, such a protocol can be built leveraging general MPC techniques.
However, this would yield an inefficient aggregation that will not be deployed in practice.
A candidate practical implementation of this SA for the specialized condition of~\Cref{eq:check} can be easily obtained by modifying the SA protocol of Bonawitz et al.~\cite{bonawitz2017practical}.
Still, we stress that this approach can also be applied to the SA protocol of Bell et al.~\cite{bell2020secure}.
In a nutshell, in~\cite{bonawitz2017practical} (and~\cite{bell2020secure}) there is an ordering over the users $\cUsers$ and each pair $(u_i,u_j)$ such that $u_i \neq u_j$ share a random secret $s_{i,j}$.
During the aggregation, each user $u_i$ masks its input $v_i$ in the following way:
\begin{align*}
	y_i & = v_i + \sum_{u_j \in \cUsers: u_i < u_j} G(s_{i,j}) - \sum_{u_j \in \cUsers: u_i > u_j} G(s_{j,i})
\end{align*}
where $G(\cdot)$ is a secure pseudorandom generator (PRG).
\iffull
Assuming no dropouts, the server can compute the aggregation $v = \sum_{u_i \in \cUsers} v_i$ from $\{y_i\}_{u_i \in \cUsers}$ as follows:
\begin{align*}
	\sum_{u_i \in \cUsers} y_i &= \sum_{u_i \in \cUsers} \left( v_i + \sum_{u_j \in \cUsers: u_i < u_j} G(s_{i,j})  - \sum_{u_j \in \cUsers: u_i > u_j} G(s_{j,i}) \right) =\\
	 & = \sum_{u_i \in \cUsers} v_i = v.
\end{align*}
\fi
\ifconference
Assuming no dropouts, the server can compute the aggregation as $v = \sum_{u_i \in \cUsers} v_i = $ $\sum_{u_i \in \cUsers} y_i$.
\fi
\iffull
To enforce the condition $C$ of~\Cref{eq:check}, we can simply substitute the PRG $G(\cdot)$ with a pseudorandom function (PRF) $F(\cdot, \cdot)$.
Now, each user $u_i$ masks its value $v_i$ by evaluating the PRF $F(s_{i,j}, \Theta^{(t)}_i)$ instead of $G(s_{i,j})$, i.e.,
\begin{align*}
	y_i & = v_i + \sum_{u_j \in \cUsers: u_i < u_j} F(s_{i,j},\Theta^{(t)}_i) - \sum_{u_j \in \cUsers: u_i > u_j} F(s_{j,i},\Theta^{(t)}_i)
\end{align*}
where $\Theta^{(t)}_i$ are the parameters received by $u_i$ from $S$.
\fi
\ifconference
To enforce the condition $C$ of~\Cref{eq:check}, we can simply substitute the PRG $G(\cdot)$ with the evaluation of a pseudorandom function (PRF) $F(s_{i,j}, \Theta^{(t)}_i)$ where $\Theta^{(t)}_i$ are the parameters received by $u_i$ from $S$.
\fi
It is easy to see that the aggregation remains hidden if two (or more) honest users receive two different parameters.
Hence, the server $S$ can not execute a model inconsistency attack.
We stress that the presented solution (as discussed in~\cite{bonawitz2017practical}) is not resilient to dropouts.
Still, the same technique \iffull(of replacing the PRG with a PRF to force the condition of~\Cref{eq:check})\fi \ can be applied seamlessly to both SA protocols (of Bonawitz et al.~\cite{bonawitz2017practical} and Bell et al.~\cite{bell2020secure}) that handle users dropouts.
This second approach does not need any additional round of communication. Moreover, unlike the previous approach (i.e., parameter validation using signatures), it preserves the communication complexity of the original SA protocols~\cite{bonawitz2017practical,bell2020secure} since we do not need to exchange signatures between active users.
% It is important to note that these solutions (\emph{parameter validation using signatures} and \emph{conditional secure aggregation}) hinge on the availability of a PKI. When this assumption is removed, patching model inconsistency becomes consistently harder due to the underlying communication topology.

Note that SA protocols such as~\cite{bonawitz2017practical,bell2020secure} (secure in the malicious setting) already require a PKI. Hence, the proposed solutions (parameter validation using signatures and conditional secure aggregation) can be implemented by using the existing PKI of the SA protocol.
Nevertheless, patching model inconsistency becomes consistently harder for variations of the vanilla FL protocol such as Asynchronous Federated Learning~\cite{nguyen2021federated, DBLP:journals/corr/abs-2109-04269} which is gaining substantial interest thanks to its practical advantages. For instance, in the asynchronous SA protocol proposed in~\cite{asyncsa}, solving model inconsistency would be a difficult task as aggregating model updates produced by different models is allowed by design. In these directions, solving model inconsistency efficiently remains an open problem.

\subsection{Differential privacy and Secure Aggregation}\label{sec:dp-mitigation}

%A more general mitigation technique would be using Differential Private-SGD algorithms~\cite{dp} which prevent gradient inversion attacks when used properly~\cite{invg1, robbing, huang2021evaluating}. However, unlike previous solutions, differential privacy (DP) comes with a high utility cost, especially in the context of FL~\cite{salvagingFL}. Note that the combination of SA and DP is still susceptible to model inconsistency attacks since the adversary can isolate the gradient of the target user; thus, parameters may need to be calibrated to ensure the aggregate noise is effective. This will be the subject of future work.

One way to protect against the proposed attacks would be to mix SA with Differential Private-SGD algorithms~\cite{dp}. {However, unlike previous solutions (Section~\ref{sec:heuristic-mitigation}), differential privacy (DP) comes with a high utility cost, especially in the context of FL~\cite{salvagingFL}.
Regardless, standard} central-DP approaches~\cite{brendan2018learning, ramaswamy2020training} are ineffective when the parameter server is malicious. Here, the DP-noise is applied only after the model updates have been aggregated by the server. Trivially, if the server is malicious, it can just skip this step and obtain the target’s gradient in clear.
Therefore, the proposed attacks remain unaffected. Peculiarly, pure central-DP is the approach used in state-of-the-art implementations~\cite{tffed} and employed in real-world deployments of FL~\cite{ramaswamy2020training, googledpftl}.  Nonetheless, user-level differential privacy can still be efficiently obtained in the presence of a dishonest parameter server by relying on the distributed-DP model~\cite{kairouz2021distributed, pmlr-v162-chen22s, NEURIPS2021_285baacb, chen2022fundamental} that combines (partial) local noise application and SA to securely simulate central-DP.\footnote{Note that the server can still exploit the $m$ compromised users allowed by the threat model and force them to participate in SA with zero gradient without applying the required noise. In the distributed DP model, this lets the server weaken the privacy guarantee for the target model update proportionally to $m$.}
More generally, as previous works have exhaustively shown it \cite{invg1, robbing, huang2021evaluating, papernot}, the application of local noise (\ie local / distributed-DP) is sufficient to prevent gradient inversion and inference attacks on users' model updates, and, therefore, the introduced attacks. In Appendix \ref{app:localDPSA}, we offer a more detailed analysis of the privacy provided by the combination of SA and local-DP against the proposed attacks.
\par

Nevertheless, the combination of SA and DP~\cite{kairouz2021distributed,StevensLWE,chen2022fundamental, pmlr-v162-chen22s, NEURIPS2021_285baacb} is still partially susceptible to model inconsistency attacks since the adversary can isolate the gradient/parameters of the target user in the aggregation. {That is, the server can still force the final aggregated value to be a function of the sole target's training set}. Thus, while a suitable amount of local noise still ensures the \TT{privacy by aggregation} property of SA\footnote{To be precise, it implies a consistently stronger form of privacy than \TT{privacy by aggregation} alone as the model update is now also differentially private.}, {the \TT{privacy by shuffling} property remains violated. Indeed, the information leaked from the aggregated model updates can still be traced back to the target---which is known to be the only source of information in the final aggregated value}.

\section{Conclusion}\label{sec:conclusion}
Our research found that federated learning implementations are susceptible to a critical vulnerability caused by incorrect usage of secure aggregation. As a result, the latter does not provide any additional security to users against a malicious server (even if a trusted PKI is assumed).
The primary reason for the security issue is the lack of parameter validation, which would have prevented the server from providing inconsistent views of the global parameters to users.

We emphasize that the proposed attacks are just representative examples of threats induced by model inconsistency, and that other attacks may be devised by exploiting the same general intuition.
In order to protect users' privacy from current and future attacks, we argue that federated learning implementations must account for model inconsistency and prevent it at its source.

\begin{acks}
	The first author was supported by \textit{Fondation Botnar}. The second author was supported by the Carlsberg Foundation under the Semper Ardens Research Project CF18-112 (BCM). The third author was supported by a grant from the Commonwealth Cyber Initiative (CCI). We acknowledge the generous support of Accenture and the collaboration with their Labs in Sophia Antipolis.
\end{acks}

\bibliographystyle{ACM-Reference-Format}
\bibliography{biblio}

\appendix
\ifconference

\section{Alternatives to federated learning}\label{sec:related-work-alternative-defenses}

\section{Security of secure aggregation}\label{sec:security-SA}

\fi

\section{Gradient suppression for arbitrary architectures}\label{sec:gsea}
In Section~\ref{sec:attack1} we focused on \relu-based layers as these are instrumental for our second attack (Section~\ref{sec:attack2}). Nevertheless, gradient suppression can be achieved for networks composed of arbitrary activation functions at the cost of flexibility and granularity.
	Intuitively, we can force any differentiable function to unconditionally produce a zero gradient by constraining it to become constant with respect to the differentiated terms. In the context of neural networks, this means making the loss function constant with respect to the trainable parameters of the network. Formally, given a neural network $f_{\Theta^{(t)}}$ and a loss function $\loss$, this can be achieved by bringing $\Theta^{(t)}$ in a state such that:
\begin{equation}
	\label{eq:fgdea}
	\forall x, y, \ \loss(y, f_{\Theta^{(t+1)}}(x)) = c,
\end{equation}
where $c$ is an arbitrary constant and $\Theta^{(t+1)}$ represents any possible alteration of the parameters of the network achievable starting from $\Theta^{(t)}$. Intuitively, when in this state, the parameters of the network have zero gradient since their alteration does not affect the loss function.

Bringing an arbitrary architecture in the state described in~\Cref{eq:fgdea} is trivial, and, for standard feedforward networks, this requires to act only on the kernels of the \textit{last and penultimate layer.} Hereafter, we refer to the composition of these last two layers as $\phi_l(\phi_p(a \otimes \theta_p + b_p)\otimes \theta_l + b_l)$ where $\phi_l$ and $\phi_p$ are the two arbitrary activation functions for the last and penultimate layer respectively and~$a$ represents the intermediate state of the network up to the penultimate layer.

To suppress the gradient for most of the network, the kernel of the last layer $\theta_l$ must be set to~$[0]$. In this way, all transformations applied to the network's input by the layers up to the last layer are nullified. Intuitively, this cuts off all parameters of the layers up to the last one from the loss computation, making their derivative~$0$ for every input. However, in this state, the kernel $\theta_l$ still receives the gradient since its alteration still affects the network's output (and so, the loss function.)

Now, to suppress the gradient for the kernel~$\theta_l$ of the last layer, the attacker needs to act on its input; that is, the output of the penultimate layer. In particular, in order to cut off the contribute of the kernel~$\theta_l$ from the loss computation, the attacker needs the output of the penultimate layer to be~$[0]$, i.e., $\forall a,\ \phi_p(a \otimes \theta_p + b_p)=[0]$.
Indeed, given the multiplicative nature of the operation~$\otimes$ between the kernel~$\theta_l$ and the $\phi_p(a \otimes \theta_p + b_p)$, the assignment $\phi_p(a \otimes \theta_p + b_p)=[0]$ completely nullifies the contribute of $\theta_l$ in the layer output, making its derivative zero.

This can be achieved by choosing appropriate values for the~$\theta_p$ and~$b_p$ in the penultimate layer. In particular, the attacker can set~$\theta_p=[0]$ and play with the bias term~$b_p$ to force the activation~$\phi_p$ to output zero. Indeed, when~$\theta_p=[0]$, we have that $\phi_p(x \otimes \theta_p + b_p) = \phi_p(b_p)$ and the attacker just needs to set $b_p$ in such a way that $\phi_p(b_p)=[0]$. This is possible for almost every activation function. In practice, this is even possible for the \textit{Sigmoid} function that is zero only when its input is $-\infty$.\footnote{It is sufficient to set $b_p$ to a large negative number.}

After having nullified the contribution of all kernels in the network, the model is now the constant function $f_\Theta(x) = \phi_l(b_l)$.
Therefore, the bias term $b_l$ (if any) can still receive a non-zero gradient during the application of SGD. Also in this case, the server can either remove the last bias term from the architecture or ignore it during the attack phase (\eg gradient inversion).
%\ifconference
%A trivial solution for the malicious server is to avoid bias terms when defining the architecture of the model.\footnote{Depending on the learning task, this can be a meaningful architectural choice.} Alternatively, the malicious server can ignore bias terms during the gradient inversion.
%Indeed, these terms represent only a tiny portion of the total number of trainable parameters of the network. For instance, in the case of a ResNet50 trained on ImageNet~\cite{resnet}, the bias vector in the final layer counts for only $4\cdot10^{-5}\%$ of the total number of parameters.
%\fi

\iffalse
\ifconference
\section{Mitigate model inconsistency using digital signatures}\label{sec:appendix-signature}
\input{mitigation/signature-approach}
\fi
\fi

\section{Canary-gradient on FedAVG}
\label{app:canary_avg}
\begin{figure}
	\centering
	
	\begin{subfigure}{.8\linewidth}
		\includegraphics[trim = 0mm 0mm 0mm 0mm, clip, width=1\linewidth]{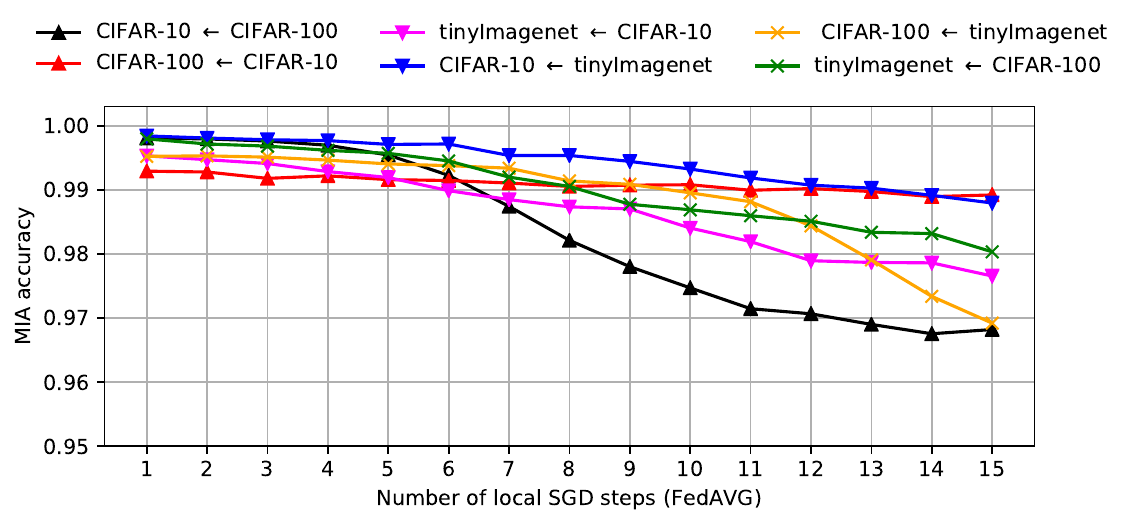}
		\caption{$lr\myeq 0.001$}
		\label{fig:miaresult_fedavg_a}
	\end{subfigure}

	\begin{subfigure}{.8\linewidth}
		\includegraphics[trim = 0mm 0mm 0mm 17mm, clip, width=1\linewidth]{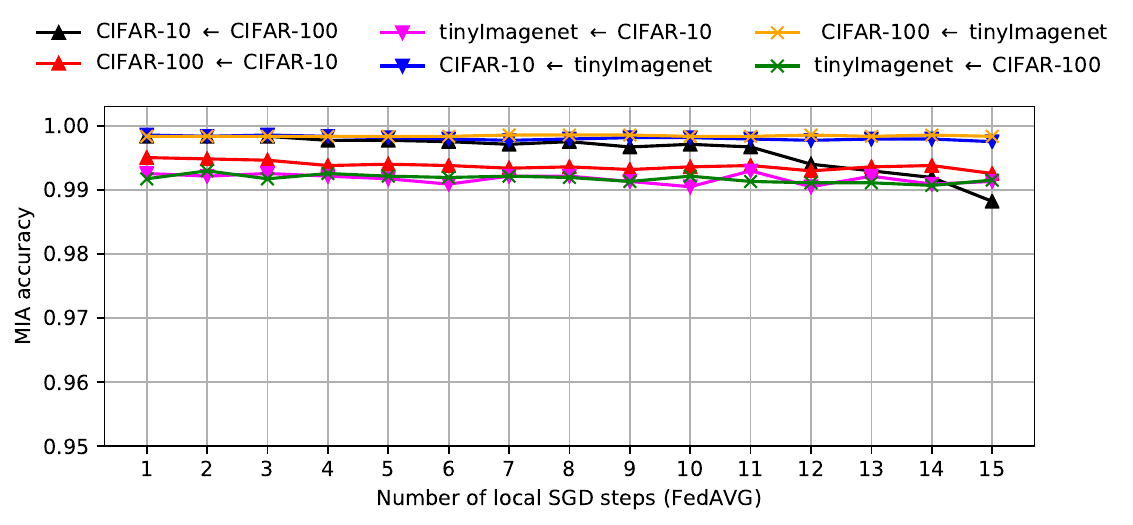}
		\caption{$lr\myeq 0.0005$}
		\label{fig:miaresult_fedavg_b}
	\end{subfigure}

	\caption{Average accuracy of canary-gradient attacks for six different setups with increasing number of FedAVG local training steps for a ResNet20.}
	\label{fig:miaresult_fedavg}

\end{figure}
At least theoretically, extending the canary-gradient attack presented in~\Cref{sec:attack2} to the FedAVG protocol can be problematic. In FedAVG, a user locally applies multiple iterations of SGD, modifying the parameters of the model. Thus, there is the possibility that the user overwrites the canary functionality during the process. However, this can be easily prevented by the server. Indeed, given the stateless nature of users in FL, the parameter server must distribute the hyper-parameters needed for the training process. In this direction, the server takes care of the learning rate. Therefore, the server can simply select a low learning rate to preserve the canary's functionality. The rationale here is that the perturbation of the local parameters induced by the local training steps is proportional to the learning rate. A low learning rate ensures to the server a bounded modification of the local parameters and, so, a limited perturbation of the hidden canary functionality originally injected in the model. Instead, the partial dead-kernel in the non-target's model cannot be revived regardless of the number of iterations and the given learning rate.
\par

 In practice, even setting a standard learning rate such as $0.001$ (which is quite common, especially towards the end of the training) seems enough to preserve the canary functionality during the local learning iterations. This is shown in~\Cref{fig:miaresult_fedavg_a}, where the canary functionality is tested after an increasing number of local training steps. We rely on the same setup of~\Cref{sec:results}, and we use a batch size of $64$. As it can be observed, the accuracy of the attack does naturally decrease with the increase in the number of iterations. However, this performance loss is limited, and the accuracy always remains in the~$97\%$-range even in the worst case. Nevertheless, there is a trade-off between effectiveness and stealthiness of the attack, and such a performance loss can be further reduced by considering a lower learning rate. This is shown in~\Cref{fig:miaresult_fedavg_b}, for a learning rate of $0.0005$. In this case, we do not need model inconsistency, and the server can send the same hyper-parameters (learning rate) to all the users in the pool without distinction.

Once computed the output of the secure aggregation, the server can state if the canary has been triggered by cheeking if $\xi_{t} \neq \xi_{t+1}$. Intuitively, the parameters $\xi$ are modified if and only if the target user produced a non-zero gradient for $\xi$ during the training at least once (\ie the trigger condition has been met).

\section{A Note on the combination of SA with Local Differential Privacy as a Defense}
\label{app:localDPSA}
When SA is combined with local-DP, the amount of noise applied to the model update recovered by our attacks is proportional to the number of users participating in the round.

In the local-DP setting without SA, the model update of the user $u_\labeltarget$ accessible by the parameter server is:

\begin{equation*}
\hat{\Delta}^{\Theta^{(t)}}_{\dataset^{(t)}_\labeltarget} + \mathfrak{N}_{\epsilon},
\end{equation*}

where $\mathfrak{N}_{\epsilon}$ is the local noise applied by $u_\labeltarget$ to make the model updated $(\epsilon)$-differentially-private and $\hat{\Delta}^{\Theta^{(t)}}_{\dataset^{(t)}_\labeltarget}$ is the clipped version of the model update.  However, the situation is different when SA is combined with local-DP. In this case, assuming that all the active users $\cUsers^{(t)}$ apply an equal amount  of noise, then the attacks recover a model update defined as:

\begin{equation}
\label{eq:saldp}
\hat{\Delta}^{\Theta^{(t)}}_{\dataset^{(t)}_\labeltarget} + \mathfrak{N}_{\epsilon} + (|\cUsers^{(t)}|-1) \cdot \mathfrak{N}_{\epsilon}.
\end{equation}

As the target model update inherits the input noise added by non-target users, the amount of noise and, therefore, the degree of protection increases with the number of active users.\\

\begin{figure}
\begin{centering}
\includegraphics[trim = 0mm 0mm 0mm 0mm, clip, width=.8\linewidth]{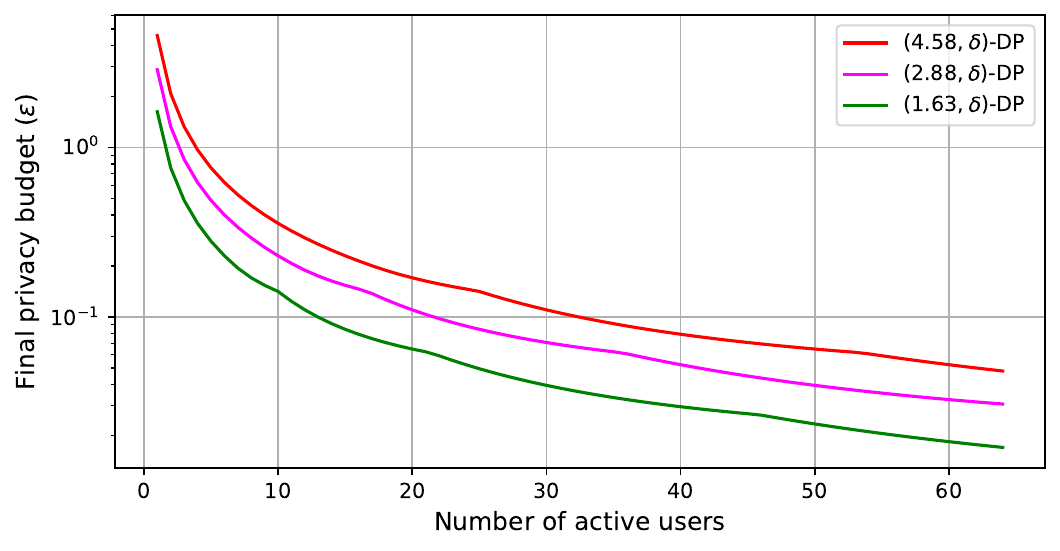}\\
\caption{Noise amplification effect on the target's model update when SA is combined with local-DP.}
\label{fig:SALDP}
\end{centering}
\end{figure}

In practice, considering DP-SGD based on the Gaussian mechanism~\cite{dp} and assuming an ideal functionality of SA working on real vectors\footnote{Current SA protocols work in the discrete domain. However, the sum of discrete (independent) Gaussians is not a discrete Gaussian (see \cite{kairouz2021distributed} for details).}, we can rewrite~\Cref{eq:saldp} as:

\begin{equation*}
\hat{\Delta}^{\Theta^{(t)}}_{\dataset^{(t)}_\labeltarget} + \mathcal{N}(0,\ |\cUsers^{(t)}| \cdot \sigma^{2}_{(\epsilon,\delta)} ),
\end{equation*}

where $\sigma^{2}_{(\epsilon,\delta)}$ is the {variance} of the Gaussian distribution required to achieve $(\epsilon,\delta)$-differential-privacy. \Cref{fig:SALDP} shows the \TT{privacy amplification effect} of the combination of SA and local-DP with respect to an increase in active users for three different settings of local-DP: $(\epsilon \myeq 4.58, \delta)$, $(\epsilon \myeq 2.88, \delta)$, and $(\epsilon \myeq 1.63, \delta)$ with $\delta \myeq 5 \cdot 10 {-5}$. The y-axis reports the final privacy budget of the model update recovered by the server after the application of SA. In the plot, we consider FedSGD with a batch size of $64$ and a local training set of $64$ instances per user. Noise multipliers are $1,\ 1.5$, and $2.5$ respectively, with $\ell_2$-norm-clip of~$1$. The privacy budget is computed with~\cite{dp} via its \texttt{tensorflow-privacy} implementation.\\

To summarize, when local-DP is combined with SA, it is possible to obtain a privacy amplification effect that is proportional to the number of non-target users. Compared to having no SA, users need to add less noise in the local-DP+SA regime to achieve the same level of protection when using a suitable trust-model.\footnote{Other users must be honest and add the expected amount of noise.}
\end{document}